%% file: paper.tex
\RequirePackage{etex}
\RequirePackage{xcolor}

\newif\ifarxiv
\arxivtrue

\ifarxiv
  \documentclass[10pt]{article}
  \usepackage[letterpaper,margin=1.5in]{geometry}
  \usepackage{natbib}
\else
  \documentclass{msc}
\fi

\RequirePackage{hyperref}
\usepackage{amsmath}
\usepackage{amsthm}
\usepackage{todonotes}
\usepackage{linguex}
\usepackage{stix2}
\usepackage{mathpartir}
\usepackage{stmaryrd}
\usepackage{comment}
\usepackage{cgloss}
\usepackage[normalem]{ulem}
\usepackage{enumitem}

\usepackage{agda/latex/agda}
\usepackage{catchfilebetweentags}
\usepackage[utf8]{inputenc}
\usepackage{newunicodechar}
\newunicodechar{∫}{\ensuremath{\int}}
\newunicodechar{Σ}{\ensuremath{\Sigma}}
\newunicodechar{→}{\ensuremath{\to}}
\newunicodechar{≡}{\ensuremath{\equiv}}
\newunicodechar{λ}{\ensuremath{\lambda}}

\usepackage[mcolblock]{leipzig}
\newleipzig{part}{part}{particle}

\theoremstyle{definition}
\newtheorem{example}{Example}[subsection]
\newtheorem{remark}[example]{Remark}
\newtheorem{notation}[example]{Notation}

\newcommand\cites[1]{\citeauthor{#1}'s\ (\citeyear{#1})}

\newcommand{\eat}{\textsf{eat}}

\newcommand{\human}{\textsf{human}}
\newcommand{\man}{\textsf{man}}
\newcommand{\woman}{\textsf{woman}}
\newcommand{\john}{\textsf{john}}
\newcommand{\mary}{\textsf{mary}}
\newcommand{\blackCat}{\textsf{blackCat}}
\newcommand{\tom}{\textsf{tom}}
\newcommand{\tomIsBlack}{\textsf{tomIsBlack}}
\newcommand{\felix}{\textsf{felix}}
\newcommand{\felixIsBlack}{\textsf{felixIsBlack}}

\newcommand{\ann}{\textsf{ann}}
\newcommand{\talk}{\textsf{talk}}
\newcommand{\meet}{\textsf{meet}}
\newcommand{\run}{\textsf{run}}
\newcommand{\die}{\textsf{die}}
\newcommand{\cat}{\textsf{cat}}

\newcommand{\apple}{\textsf{apple}}
\newcommand{\balloon}{\textsf{balloon}}
\newcommand{\pop}{\textsf{pop}}
\newcommand{\popped}{\textsf{popped}}
\newcommand{\blk}{\textsf{black}}
\newcommand{\striped}{\textsf{striped}}
\newcommand{\book}{\textsf{book}}
\newcommand{\quick}{\textsf{quick}}
\newcommand{\mouse}{\textsf{mouse}}

\newcommand{\butter}{\textsf{butter}}
\newcommand{\toast}{\textsf{toast}}

\newcommand{\withknife}{\textsf{with}\_ \textsf{knife}}
\newcommand{\atmidnight}{\textsf{at}\_ \textsf{midnight}}
\newcommand{\inbathroom}{\textsf{in}\_ \textsf{bathroom}}

\newcommand{\dog}{\textsf{dog}}
\newcommand{\bark}{\textsf{bark}}

\newcommand{\quantity}{\textsf{quantity}}
\newcommand{\height}{\textsf{height}}

\newcommand{\weight}{\textsf{weight}}
\newcommand{\width}{\textsf{width}}
\newcommand{\vvolume}{\textsf{volume}}
\newcommand{\Entity}{\textbf{Entity}}
\newcommand{\Occ}{\textbf{Occ}}
\newcommand{\water}{\textsf{water}}

\newcommand{\natunits}{\textsf{nu}}
\newcommand{\liter}{\textsf{liter}}
\newcommand{\meter}{\textsf{meter}}
\newcommand{\kilogram}{\textsf{kilogram}}
\newcommand{\several}{\textsf{several}}

\newcommand{\B}{\textsf{B}}
\newcommand{\U}{\textsf{U}}
\newcommand{\A}{\textsf{A}}
\newcommand{\Patient}{\textsf{Und}}

\newcommand{\sumNP}{\textstyle\sum^\NP}
\newcommand{\sumEvt}{\textstyle\sum^\Evt}

\newcommand{\Evt}{\textbf{Evt}}
\newcommand{\Tel}{\textbf{Tel}}
\newcommand{\Atel}{\textbf{Atel}}
\newcommand{\Cul}{\textbf{Cul}}

\newcommand{\Actor}{\textbf{Act}}
\newcommand{\actTm}{\textbf{act}}
\newcommand{\Undergoer}{\textbf{Und}}
\newcommand{\undTm}{\textbf{und}}
\newcommand{\Vect}{\textsf{Vec}}
\newcommand{\type}{\textsf{ type}}
\newcommand{\Prop}{\textbf{Prop}}
\newcommand{\NP}{\textbf{NP}}
\newcommand{\CN}{\textbf{CN}}
\newcommand{\Boundedness}{\textbf{Bd}}
\newcommand{\IntAdj}{\textbf{IntAdj}}
\newcommand{\IA}{\textbf{IA}}
\newcommand{\El}{\textbf{El}}
\renewcommand{\Lift}{\textbf{Lift}}
\newcommand{\SumIsCount}{\Sigma\textbf{IsCount}}
\newcommand{\IANPIsNP}{\textbf{IANPIsNP}}
\newcommand{\IARespectsIsA}{\textbf{IARespectsIsA}}
\newcommand{\oplusPreservesIA}{\oplus\textbf{PreservesIA}}
\newcommand{\EvtEntIsNP}{\textbf{EvtEntIsNP}}
\newcommand{\EvtAmtIsNP}{\textbf{EvtAmtIsNP}}
\newcommand{\NPIsOneNP}{\textbf{NPIsOneNP}}
\newcommand{\OneNPIsNP}{\textbf{OneNPIsNP}}
\newcommand{\Degree}{\textbf{Degree}}
\newcommand{\Units}{\textbf{Units}}
\newcommand{\Nat}{\textbf{Nat}}

\newcommand{\AmountOf}{\textbf{AmountOf}}
\newcommand{\Result}{\textbf{Result}}
\newcommand{\isCul}{\textbf{isCul}}
\newcommand{\isA}{\textbf{isA}}
\newcommand{\isArefl}{\textbf{isArefl}}
\newcommand{\isAtrans}{\textbf{isAtrans}}
\newcommand{\isCount}{\textbf{isCount}}
\newcommand{\State}{\textbf{State}}
\newcommand{\Prff}{\textbf{Prf}}
\newcommand{\prf}{\textsf{prf}}

\newcommand{\Univ}{\textbf{Type}}
\newcommand{\CulOrAtel}{\textbf{CulOrAtel}}

\newcommand{\Activity}{\textbf{Activity}}
\newcommand{\Accomplishment}{\textbf{Accomplishment}}
\newcommand{\Achievement}{\textbf{Achievement}}
\newcommand{\EventNucleus}{\textbf{Event nucleus}}
\newcommand{\Culminate}{\textbf{Culminate}}
\newcommand{\Consequent}{\textbf{Consequent}}

\newcommand{\fst}{\textbf{fst}}
\newcommand{\snd}{\textbf{snd}}
\newcommand{\var}[1]{\textit{#1}} 
\newcommand{\semantics}[1]{\llbracket #1 \rrbracket}

\begin{document}

\title{A dependently-typed calculus of event telicity and culminativity}

\ifarxiv
  \author{Pavel Kovalev \and Carlo Angiuli}
  \date{June 2025}
\fi

\ifarxiv\maketitle\fi

\begin{abstract}
We present a dependently-typed cross-linguistic framework for analyzing the telicity and
culminativity of events, accompanied by examples of using our framework to model English sentences.
Our framework consists of two parts. In the nominal domain, we model the boundedness of noun phrases
and its relationship to subtyping, delimited quantities, and adjectival modification. In the verbal
domain we define a dependent event calculus, modeling telic events as those whose undergoer is
bounded, culminating events as telic events that achieve their inherent endpoint, and consider
adverbial modification. In both domains we pay particular attention to associated entailments. Our
framework is defined as an extension of intensional Martin-L\"of dependent type theory, and the
rules and examples in this paper have been formalized in the Agda proof assistant.
\end{abstract}

\section{Introduction}\label{sec:Introduction}
\input{sections/introduction}

\section{Type-theoretic background}\label{sec:Background}
\input{sections/background}

\section{The nominal domain}\label{sec:Nominal}
\input{sections/nominal}

\section{The verbal domain}\label{sec:Verbal}
\input{sections/verbal}

\section{Agda implementation}\label{sec:Agda}
\input{sections/agda}

\section{Related and future work}\label{sec:Future}
\input{sections/future}

\bibliographystyle{msclike}
\bibliography{refs, pavel}

\end{document}

%% file: sections/introduction.tex
Since the inception of the field of formal semantics of natural language
\citep{1970_Montague_EngAsFormLang, 1970_Montague_UnivGram, 1973_Montague_Proper}, Church's simple
type theory \citep{1940_Church} and its variations have been dominant in formal natural language
analysis \citep{1990_Gamut_BOOK}. More recently, however, there is growing interest in using
\emph{dependent type theory} \citep{MartinLof75itt} for this purpose \citep{Sutton2024}. Unlike
simple type theory, dependent type theory allows greater expressivity at the type level by allowing
types to be indexed by terms, making it possible to analyze or model certain natural language
phenomena that are problematic or impossible in the setting of simple type theory
\citep{ChatzikyriakidisCooper2018, 2020_Chatzikyriakidis_BOOK, Sutton2024, chatzikyriakidis2025types}.

In this paper, we develop a framework within dependent type theory for modeling fragments of natural
language involving the phenomena of \emph{telicity}, the property of an event having an inherent
endpoint, \emph{culminativity}, the property of a telic event attaining its endpoint, and their
counterparts \emph{atelicity} and \emph{non-culminativity}.\footnote{These phenomena lie in the
domain of inner aspect (also known as lexical aspect, or situation aspect), as opposed to the domain
of outer aspect (also known as grammatical aspect, or viewpoint aspect); see e.g.\
\cite{1997_Smith_BOOK}. We will not be concerned with outer aspect, nor will we be concerned with
the tense.} As widely discussed in the linguistic literature \citep{1972_Verkuyl_BOOK,
1993_Verkuyl_BOOK, 1979_Dowty_BOOK, tenny1987grammaticalizing, 1994_Tenny_BOOK, Moens1988,
1989_Krifka_CONF, 1992_Krifka, 1998_Krifka, 1991_Jackendoff, 1996_Jackendoff}, for a large class of
verbs, the telicity of an event depends on the properties of its undergoer, the most patient-like
argument of the verb describing the event.

Our framework makes essential use of type dependency not only to account compositionally for the
relationship between the telicity of events and the properties of their undergoers, but also to
account for the fact that the property of (non-)culminativity is only defined for telic events,
because (non-)culminativity presupposes the existence of an inherent endpoint.

\subsection{Telicity and culminativity}\label{sec:bg-telicity}

A verbal predicate is \emph{telic} if the eventuality it characterizes has an inherent endpoint.  We
will speak of a \emph{telic event} to mean an eventuality characterized by a telic verbal
predicate.%
\footnote{In our framework, we distinguish between \emph{events} and their \emph{occurrences}. An
\emph{event} is the semantic object introduced by a verbal predicate, and it can have (possibly
many) \emph{occurrences}, which are particular concrete realizations. Thus, when we use the phrase
\emph{telic event}, the telicity is attributed to what the predicate characterizes (the event in
this sense), not to any individual occurrence.}
Although it is difficult to precisely define inherent endpoints, it is relatively easy to
illustrate what does and does not constitute an inherent endpoint, and what it means for that
endpoint to be attained.

Every sentence in \ref{inherent endpoint} below describes an event with an inherent endpoint. (By
abuse of language, we henceforth conflate sentences describing events with the described events.)
The endpoint is lexically specified in each sentence either as an argument of the verb, as in
\ref{John ate the soup} -- \ref{Three balloons popped},\footnote{Such an argument is referred to as
an \emph{object of termination} in \cite{Voorst1988}. In our setting, a more proper name would be an
\emph{object of culmination}, where ``object'' is used in a non-syntactic sense.} or as an adjunct,
as in \ref{John walked to the park}. For example, in \ref{John ate the soup}, \emph{the soup}
supplies an inherent endpoint, and we say that this endpoint is attained if the soup is completely
consumed.\footnote{To emphasize, \emph{the soup} only supplies an inherent endpoint, but it would
probably be counterintuitive to say that \emph{the soup} is itself an inherent endpoint.} In other
words, any event with an inherent endpoint has some resulting state that is naturally associated
with it, and we say that this endpoint is attained if the resulting state ensues. The resulting
states associated with the sentences in \ref{inherent endpoint} can be expressed by the sentences in
\ref{associated states}.

\begin{center}%
\begin{minipage}{0.5\textwidth}
\ex. \label{inherent endpoint}
\a. John ate the soup.\label{John ate the soup}
\b. John repaired my computer. \label{John repaired my computer}
\b. John wiped the table. \label{John polished the vase}
\b. Three balloons popped. \label{Three balloons popped}
\b. John walked to the park. \label{John walked to the park}

\end{minipage}%
\begin{minipage}{0.5\textwidth}
\ex. \label{associated states}
\a. There is no soup left.
\b. My computer is repaired.
\c. The table is clean.
\d. Three balloons are popped.
\e. John is in the park.

\end{minipage}%
\end{center}

In contrast, the sentences in \ref{no inherent endpoint} below have no inherent endpoints. For
example, the sentence \ref{John ate soup} has no inherent endpoint because it can potentially
continue indefinitely; unlike \ref{John ate the soup}, there is no point at which the event of
eating soup would naturally stop.\footnote{One might object that it cannot continue indefinitely
because it is in the past tense, but the contrast that we intend to convey is still present if one
replaces the past tense with the future tense.} Unlike \emph{the soup} in \ref{John ate the soup},
\emph{soup} in \ref{John ate soup} does not supply an inherent endpoint, and as a result, there does
not exist any intended resulting state associated to \ref{John ate soup}.

\ex. \label{no inherent endpoint}
\a. John ate soup. \label{John ate soup}
\b. John repaired computers.
\b. John wiped tables. \label{John polished vases}
\b. Balloons popped.\label{Balloons popped}
\b. John walked in the park.

(Note that although the sentences in \ref{no inherent endpoint} have no natural notion of (intended)
resulting state, they do all nevertheless involve some change of state due to their dynamic nature.)

An event (or more generally eventuality; see Remark~\ref{rem:eventuality}) is telic if and only if
it has an inherent endpoint; otherwise, it is atelic. Thus the events in \ref{inherent endpoint} are
telic, and the events in \ref{no inherent endpoint} are atelic. The main feature which distinguishes
telic and atelic events is that telic events have the capacity to \emph{culminate}, or attain their
inherent endpoint. In contrast, the question of whether or not an atelic event culminated is not
sensible because there is no inherent endpoint with respect to which culmination can take place.

\begin{remark}\label{rem:eventuality}
Following \cite{1981_Bach, 1986_Bach}, we use the term ``eventuality'' in a very broad sense,
similarly to the term ``state of affairs'' in \cite{1997_Valin_BOOK}, subsuming the four classical
Aktionsart classes of \citet{1957_Vendler, 1967_Vendler_BOOK}, namely states, activities,
achievements, and accomplishments.
It is common to divide eventualities into states and non-states and, among non-states, to distinguish processes---Vendlerian activities---and events---Vendlerian achievements and accomplishments---see \cite{1978_Mourelatos, 1986_Bach}. To keep the metalanguage lighter, we adopt the terminological convention that ``event'' is a cover term for non-static eventualities (thus including both processes and events in the narrower sense). This convention is purely terminological, and it is not meant to collapse the process--event distinction. Our analysis of telicity will be restricted to the class of verbal predicates whose telicity varies with the properties of an argument (rather than being induced by adjuncts); from this perspective, purely process predicates that we mention in the paper, such as \emph{run} or \emph{drive a car}, function mainly as atelic points of comparison.
	
	A brief comment is in order regarding the potentially distinct fifth Aktionsart class of semelfactives, such as \emph{cough}, \emph{blink}, or \emph{knock} \citep{1997_Smith_BOOK}.
	There is no agreement in the literature on how semelfactives should be situated relative to the classical classes, and relatedly on whether they should be classified as telic or atelic.
	One reason is that different traditions connect these labels to different diagnostics: for example, \cite{1978_Mourelatos} treats semelfactives as events in his sense on the grounds of countability (one can say \emph{blink twice}), whereas \cite{1997_Smith_BOOK} treats semelfactives as a separate class, characterized by instantaneousness and the absence of an inherent resulting state (e.g.\ there is no natural resulting state that ensues upon John knocking on the door).
	Since we are only concerned with the telicity--atelicity distinction and not with the distinction between particular Aktionsart classes, the only thing that matters for us is that semelfactives are atelic according to our definition, where telicity is understood in terms of an inherent endpoint associated with an intended resulting state, rather than countability.
	Accordingly, semelfactives count as atelic in our setting, since they do not have a natural resulting state associated with them.
\end{remark}

An event attains its inherent endpoint (i.e., culminates) if that event entails its associated
intended resulting state. In English, events expressed by most non-progressive verb phrases with a
specified inherent endpoint automatically culminate. For example, the events in \ref{inherent
endpoint} (with the possible exception of \ref{John polished the vase}, as discussed below) attain
their inherent endpoints and thus culminate. We illustrate this in \ref{Eng verbs entail culm}: with
the exception of \ref{John polished the vase but there are still}, the first clause of each sentence
entails culmination with respect to its endpoint, and for that reason the second clauses introduce
contradictory information.

\ex.\label{Eng verbs entail culm}
\a. \#John ate the soup, and I finished it.
\b. \#John repaired my computer, but the computer still doesn't work.
\b. John wiped the table, but there are still water drops on it. \label{John polished the vase but there are still}
\b. \#Three balloons popped, but one of them is still inflated.
\b. \#John walked to the park, but he never arrived there.

As for \ref{John polished the vase but there are still}, this is about as close as we can get in
English to a telic but non-culminating event. The verb \emph{wipe} is arguably ambiguous between at
least two readings. On one reading of \emph{wipe}, it denotes an activity—the mere motion of moving
a cloth back and forth—with no inherent endpoint; hence, under this reading, \ref{John polished the
vase but there are still} is atelic. On instead a result-oriented reading of \emph{wipe}, the first
clause in \ref{John polished the vase but there are still} denotes a telic event whose inherent
endpoint is that the entire relevant surface has been wiped (at least once). In ordinary discourse,
this telic reading often gives rise to a pragmatic inference that the table is clean or dry, but
that inference is not entailed by the verb itself. In \ref{John polished the vase but there are
still}, the continuation denies the expected result state (“there are still water drops”), showing
that this is an approximation of a telic but non-culminating event: the inherent endpoint is
defined, but the culmination condition fails to hold. This is, however, not a true telic
non-culminating event, since dryness is not entailed by \emph{wipe}.

 Cross-linguistically, there are languages in which non-culminating events (often termed \emph{non-culminating accomplishments}) are more robust and prominent than in English.
 We restrict ourselves with just one example
from Mandarin Chinese, which is given in \ref{mandarin nonculm}; the English translation is semantically
unacceptable in English, but the original sentence is acceptable in Mandarin.

\exg.
Mǎlì chī-le yī-gè píngguǒ, hái shèng liǎng-kǒu, bèi wǒ chī-le.
\\
Mary eat-\Pfv{} one-\Clf{} apple still left two-bite \Pass{} I eat-\Pfv{}\\ \label{mandarin nonculm}\hfill 
\glt `Mary ate an apple. There was some left over and I ate it.'
\hfill \cite[466]{2022_Gu}

\noindent
The first clause \emph{chī-le yī-gè píngguǒ} `ate an apple' is telic (it has a natural
endpoint: the apple's being fully consumed), while the follow-up \emph{hái shèng liǎng-kǒu}
`there were two bites left' denies that endpoint. The particle \emph{-le} marks event
occurrence/completion but does not force culmination, permitting a genuinely telic but
non-culminated reading.

\relax

Finally, we note that there is a class of verbs including \emph{push}, \emph{stroke}, and
\emph{drive}, whose lexical semantics does not involve an approach towards any goal, unlike verbs
such as \emph{repair}, \emph{pop}, and \emph{melt}, whose lexical semantics does involve an approach
towards some goal. Arguments of verbs like \emph{push}, \emph{stroke}, or \emph{drive} can never
introduce inherent endpoints; only adjuncts can. This is illustrated in \ref{ex:ateltotelwithcompl}.

\ex. \label{ex:ateltotelwithcompl}
\a. John drove a car. \hfill (atelic)
\b. John drove a car \emph{to Bloomington}. \hfill (telic)

\relax

To sum up, events can be telic or atelic. Telic events may be culminating or non-culminating,
whereas the status of (non-)culminativity is undefined for atelic events. In events described by
verbs whose lexical semantics encode an aim toward some goal, telicity may arise either from an
argument of the verb or from an adjunct. In contrast, events described by verbs without an aim
toward any goal are always atelic in the absence of extra linguistic content, although adding extra linguistic content such as adjuncts to the predicate describing them may render the event telic. (We
ignore such cases in this paper, leaving them to future work.)

\subsection{Relationship between telicity and undergoers}
\label{sec:Relationship between telicity and undergoers}

Following Role and Reference Grammar \citep{1997_Valin_BOOK, Bentley2023}, we use the terms
\emph{actor} and \emph{undergoer} to label the two arguments of a transitive verb. Actors and
undergoers are semantic macroroles: actors (resp., undergoers) represent the most agent-like (resp.,
patient-like) participants of an event. Actors are more general than agents; they do not have to be
sentient, and include other agent-like roles, such as instruments or forces. Our notion of undergoer
also encompasses a variety of traditional semantic roles, such as patient, theme, experiencer,
holder of a state, and similar roles. For example, the grammatical subjects in all the transitive
sentences in \ref{inherent endpoint} and \ref{no inherent endpoint} are actors, and the grammatical
objects are undergoers; the grammatical subjects in the intransitive sentences in \ref{associated
states} are all undergoers.


As discussed in Section~\ref{sec:bg-telicity}, for verbs whose lexical semantics encodes an aim to
achieve a goal, an inherent endpoint may be supplied either by an argument of the verb or by an
adjunct. Restricting attention to the former case and verbs with at most two arguments, linguists
have observed that the property of telicity in the verbal domain correlates with a certain property
of the undergoer in the nominal domain \citep{1972_Verkuyl_BOOK, 1993_Verkuyl_BOOK, 1979_Dowty_BOOK,
tenny1987grammaticalizing, 1994_Tenny_BOOK, 1989_Krifka_CONF, 1992_Krifka, 1998_Krifka,
1991_Jackendoff, 1996_Jackendoff}.

\cite{1991_Jackendoff, 1996_Jackendoff} refers to the latter property as \emph{boundedness}. A
bounded noun phrase, according to Jackendoff, is one that describes a delimited quantity of some
thing or a substance, such as \emph{three apples, five liters of water, several books, a tomato}.
Examples of noun phrases that are not bounded, or \emph{unbounded}, include \emph{apples, water,
custard}.

We can detect the correlation between telicity and boundedness by the \emph{in}-adverbial test: only
telic events can be followed by temporal \emph{in}-adverbials. Indeed, temporal \emph{in}-adverbials
specify the time span within which something has happened, and this ``happening'' can occur only if
there is an inherent endpoint to begin with; cf.\ \citet[223]{Peck2016}. On the other hand, in
English, only sentences with bounded undergoers are compatible with \emph{in}-adverbials, as the
contrast in \ref{ex:TelicityCorrelatesWithBoundedness} shows. This illustrates the correlation
between the telicity of an event and the boundedness of its undergoer.

\ex. \label{ex:TelicityCorrelatesWithBoundedness}
\a. John ate three apples in ten minutes.\label{ex:TelicityCorrelatesWithBoundednessThreeApples}
\b. $\ast$ John ate apples in ten minutes. \label{ex:TelicityCorrelatesWithBoundednessApples}

\begin{remark}\label{rem:atleast3apples}
There are some known exceptions, notably with noun phrases involving quantifiers. For example, the
noun phrase \emph{at least three apples} is not delimited in any sense, but it ought to be bounded
according to the \emph{in}-adverbial test. On the other hand, the noun phrase \emph{some apples}
seems to be delimited, but it behaves like an unbounded noun phrase with respect to the
\emph{in}-adverbial test.
\end{remark}

In English, bounded and unbounded noun phrases are syntactically marked: the unbounded noun phrases
are the bare noun phrases (i.e., noun phrases expressed by bare nouns, possibly modified by
adjectives, but without any determiners), whereas the bounded noun phrases are the non-bare noun
phrases.\footnote{As already observed in Remark~\ref{rem:atleast3apples}, noun phrases like
\emph{some apples} are exceptions to this generalization.} In article-less languages, however, bare
noun phrases \emph{can} have a definite reading.

To account for this cross-linguistic variation, we redefine the notion of boundedness so as to not
refer to the syntactic structure of a noun phrase. Under our definition, a noun phrase is bounded if
it is capable of supplying an inherent endpoint, and unbounded otherwise. That is, we treat
boundedness (resp., unboundedness) as the nominal counterpart of telicity (resp., atelicity), in the
sense that an event is telic if and only if its undergoer is bounded. (Of course, this is only
adequate for verbs whose lexical semantics encodes an aim to achieve a goal.) In
\cite{kovalev2024modeling}, the first author argues that this definition, together with appropriate
diagnostics for detecting bounded and unbounded undergoers, is an adequate generalization of
\cites{1996_Jackendoff} notion of boundedness to at least some languages that lack articles, such as
Russian, and that the two notions of boundedness agree when restricted to English.

In this paper we will restrict our attention to transitive and intransitive (but not ditransitive)
verbs whose lexical semantics encode an aim to achieve a goal (i.e., those verbs which some may call
``non-activity'' verbs), and on the cases in which telicity is induced by a verbal argument as
opposed to an adjunct. These are the verbs for which the telicity of an event correlates to the
boundedness of its undergoer.

As discussed above, we will treat the boundedness of a noun phrase as a primitive notion. In
English, the bounded noun phrases will be the non-bare noun phrases (i.e., those containing
determiners). As for languages where bare noun phrases have both bounded and unbounded readings,
such as Russian, the first author has argued in \cite{kovalev2024modeling} that bounded readings of
bare noun phrases in Russian can be interpreted either as definite descriptions or as having an
unspecified numerical quantifier. We restrict ourselves to the latter interpretation for the
purposes of this paper, leaving the modeling of definite descriptions (and more generally,
inter-clausal relationships) to future work.


\subsection{Comparison to prior work on telicity and culminativity}

Although telicity and culminativity are well-studied in the linguistic literature, previous studies
focus almost exclusively on explaining the origins of these phenomena. In comparison, our focus is
on the entailment relations associated with these notions. One can argue that the notion of
entailment is a central and the most important notion in all of semantics
\citep{1970_Montague_UnivGram, 2015_Moss_CONF}; indeed it is what distinguishes semantics from the
adjacent field of pragmatics, whose central notion is that of implicature. Despite this, prior work
has not focused on formally modeling entailment relations arising in the context of telicity. Nor
are we aware of studies which use dependent types to model telicity; the closest work we are aware
of is \cites{2020_Corfield_BOOK} sketch of the modeling of Vendler's Aktionsart classes
\citep{1957_Vendler, 1967_Vendler_BOOK}.

The vast majority of current formal analyses of telicity and culminativity can be broadly
categorized as following either the mereological approach, involving parts of events and entities,
or the lexico-semantical approach, involving predicate decomposition. The former goes back to
\cite{1986_Krifka_DISS, 1989_Krifka_CONF, 1992_Krifka, 1998_Krifka, 2001_Krifka}, while the latter
was originally proposed in \cite{1979_Dowty_BOOK} and further developed in \cite{1997_Valin_BOOK,
2005_Valin_BOOK, Bentley2023}, with \cite{1998_Malka} providing a close alternative. The two
approaches were combined in \cite{2004_Rothstein_BOOK}.

To oversimplify, the mereological approach assumes that an event comes to an end if and only if the
undergoer of the event is completely ``used up.'' For example, the event \emph{John ate three
apples} comes to an end if and only if the three apples are completely consumed. This approach
conflates telicity and culminativity; as we have discussed in Section~\ref{sec:bg-telicity}, such a
conflation is generally suitable for modeling telicity in English, but---unlike our approach---is
not straightforwardly suitable to languages with telic but non-culminating events.

On the other hand, approaches based on lexical decomposition encode the relevant inner aspectual
properties of events in the symbolic representation of the predicate describing that event. To
illustrate, \ref{ex:RRG1} and \ref{ex:RRG2} provide examples of certain kinds of lexical
representations in Role and Reference Grammar \cite[99]{Bentley2023}.

\ex. \label{ex:RRG1}
\a. The ship sank.
\b. $\text{BECOME } \textbf{sunken}\prime(\text{ship})$

\ex. \label{ex:RRG2}
\a. The submarine sank the ship.
\b. $[\textbf{do}\prime(\text{submarine},\varnothing)] \text{ CAUSE } [\text{BECOME } \textbf{sunken}\prime(\text{ship})]$

\relax

Our main critique of such representations is that they are typically not accompanied with a formal
semantics (be it model-theoretic, proof-theoretic, or otherwise). Thus although they provide a
suitable symbolic representation for sentences, the absence of a semantics makes it impossible to
discuss entailments (and consequently to perform natural language inference).

We note that---like the mereological approaches described earlier---the examples from Role and
Reference Grammar mentioned above conflate telicity with culminativity. There have been attempts to
unify the two approaches and add a separate predicate governing culminativity, notably by
\cite{2004_Rothstein_BOOK}. Although these works attempt to define the semantics, they appear to be
informal in many aspects and to conflate syntax and semantics, making them unsuitable for
formalization in the context of automated natural language inference.

In this paper we aim to rectify these issues. In particular, we distinguish the notions of telicity
and culminativity, noting further that (non-)culminativity is a property only defined for telic
events. Our approach to modeling telicity can be seen as an improvement of lexical decomposition
approaches, in which we work within (dependent) type theory instead of creating a separate
representation language.

Using type theory provides us with several benefits. First, type theory is a rich enough framework
to model the relevant aspects of telicity and culminativity and their relationship to the argument
structure in a way that doesn't conflate the two notions. Second, there is a general notion of a
model of dependent type theory (e.g., as induced by the theory of generalized algebraic theories
\citep{Cartmell86,Uemura21}), allowing us to equip our framework with a semantics as a
straightforward extension of the standard set-theoretic semantics of type theory \citep{Hofmann97}.
(See Section~\ref{sec:Future} for further discussion of this point.)

Finally, type theory has many computer implementations in modern proof assistants such as Agda
\citep{Agda}, Rocq/Coq \citep{Coq2010}, Lean \citep{DeMouraUllrich21}, and Lego/Plastic
\citep{Luo1992, 2001_Callaghan}. As a result, the framework we develop in this paper is amenable to
implementation in proof assistants; indeed, in Section~\ref{sec:Agda} we describe an Agda
implementation of all the rules and examples in this paper. Such implementations can in turn be used
for automated natural language inference or, more generally, natural language reasoning
\citep{2016_Chatzikyriakidis_CONF}. Fragments of English have previously been implemented in
Rocq/Coq \citep{2014_Chatzikyriakidis, 2011_Luo_CONF}, Plastic \citep{2012_Xue_CONF}, and Agda
\citep{Domanov2024}.

\subsection{Prior use of dependent types for language modeling}\label{sec:MTTsemantics}

The use of dependent types for natural language analysis goes back at least as far as
\cite{monnich1985untersuchungen} and \cite{1986_Sundholm_CONF}, in which they were used to model
anaphoric relationships in so-called donkey sentences \citep{1962_Geach_BOOK}. Although the first
comprehensive work on using dependent types for natural language analysis appeared around the same
time \citep{1994_Ranta_BOOK}, relatively few studies followed in the subsequent years
\citep{Boldini1997, 2000_Boldini, 2001_Boldini_CONF, Krahmer1999, Piwek2000, 2001_Ahn_DISS}.

Over the last ten to fifteen years, several lines of work have emerged applying dependent types in
the context of natural language. Perhaps the most fully developed framework among these is the
so-called MTT-semantics \citep{2009_Luo_CONF, 2012_Luo, 2020_Chatzikyriakidis_BOOK}. Another
developed line of work is Dependent Type Semantics \citep{2014_Bekki_CONF, 2017_Bekki_CONF,
Bekki2023, Matsuoka2024}, Type Theory with Records \citep{2005_Cooper, Cooper2011, 2015_Cooper_CONF,
2017_Cooper_CONF, 2023_Cooper_BOOK}; there are many other isolated works as well
\citep{2010_Dapoigny, 2011_Asher_BOOK, asher2013formalization, 2014_Retore, Grudzinska2014,
Grudzinska2017, Grudzinska2019, Grudzinska2020, 2018_Mery_CONF, 2019_OrtegaAndres, Zawadowski2024}.


In addition to the general virtues of dependent type theory mentioned above, many researchers are
turning to dependent type theory for natural language analysis because in comparison to Church's
simple type theory, it provides a more elegant account of---or in some cases, allows any account
whatsoever of---natural language phenomena including anaphoric relationships, copredication,
selectional restriction, the event quantification problem, and adverbial modification, among others
\citep{ChatzikyriakidisCooper2018, 2020_Chatzikyriakidis_BOOK, Sutton2024, chatzikyriakidis2025types}.

Our framework in this paper is inspired in many ways by MTT-semantics. We note that the term
``MTT-semantics'' stands for ``formal semantics in modern type theories'', and in principle is meant
to cover any kind of formal semantics that employs a modern (dependent) type theory. In practice,
however, this term seems to refer to the specific framework developed by \cite{2009_Luo_CONF,
2012_Luo, 2020_Chatzikyriakidis_BOOK}, whose underlying type theory is the Unifying Theory of
Dependent Types \citep{1994_Luo_BOOK}, and it is in this sense that we use the term.

The principal commonality between our work and MTT-semantics is the use of the \emph{common nouns as
types} paradigm \citep{2012_Luo_CONF, 2012_Luo}, which goes back to earlier type-theoretic works
\citep{monnich1985untersuchungen, 1986_Sundholm_CONF, 1994_Ranta_BOOK}, but stands in opposition to
the treatment in Montague semantics of common nouns as predicates. In MTT-semantics, modeling common
nouns as the types of their instances enables the modeling of linguistic phenomena that involve
subtyping, such as copredication. In this paper, it makes our treatments of the nominal and verbal
domains more uniform.\footnote{Alternative approaches include treating common nouns as setoids
\citep{Chatzikyriakidis2017, 2018_Chatzikyriakidis}, predicates \citep{Krahmer1999, 2014_Bekki_CONF,
Tanaka2017, 2017_Bekki_CONF}, and as both predicates and types \citep{2014_Retore,
2021_Babonnaud_CONF}. See \cite{2017_Chatzikyriakidis_CONF} for a comparison between the common
nouns as types and as predicates approaches, and \citet[Section~2.6]{2020_Corfield_BOOK} for further
discussion.}

We also adopt a variant of the dependent event types of MTT-semantics \citep{2017_Luo_CONF},
although our approach is different in that our dependent event types are collections of events that
are themselves types. As we will discuss in Section~\ref{sec:Verbal}, this modification allows us to
apply adverbial modification in parallel with adjectival modification while also keeping track of
the telicity and culminativity status of events.

Our biggest departure from MTT-semantics is the underlying type theory itself. MTT-semantics is
built on the impredicative Unifying Theory of Dependent Types \citep{1994_Luo_BOOK} extended by
numerous features such as coercive subtyping \citep{Luo1997, 1999_Luo, 2013_Luo}, subtype universes
\citep{2021_Maclean_CONF, Bradley2023}, and dot-types \citep{2009_Luo_CONF, 2011_Luo_CONF, 2012_Luo,
2015_Chatzikyriakidis_CONF, Chatzikyriakidis2017, 2018_Chatzikyriakidis}. These features play key
roles in modeling various natural language phenomena; for example, dot-types were introduced to
handle copredication, subtype universes to handle gradable adjectives, and weak sums (which are
definable using impredicativity) are used to account for the donkey anaphora \citep{2021_Luo}.

In contrast, we work inside predicative intensional Martin-L\"of type theory \citep{MartinLof75itt},
a very well-understood system from both the theoretical and practical standpoint. In part this is
because we are not attempting to model all the linguistic phenomena considered by MTT-semantics.  In
other cases, we make a concerted effort to avoid sophisticated features; for example, in
Section~\ref{sec:Subtyping in the nominal domain} we model subset inclusions not using coercive
subtyping as in MTT-semantics but rather in terms of explicit inclusion functions.

Working inside a more standard dependent type theory has several benefits. First, it allows us to
formalize our framework inside the Agda proof assistant with minimal modifications. Second, by
making fewer background assumptions, our work is directly compatible with more extensions of type
theory, including homotopy type theory \citep{HoTT13}. Most importantly, the set-theoretic (and
category-theoretic and even homotopy-theoretic) semantics of Martin-L\"of type theory are
well-studied \citep{Hofmann97}, in contrast to the model theory of impredicative type theories with
coercive subtyping, which is much less clear. The latter has led to several papers attempting to
clarify that MTT-semantics is model-theoretic in addition to being proof-theoretic
\citep{2014_Luo_CONF, LuoBOTH2019}.

\subsection{Contributions}

We develop a cross-linguistic framework within Martin-L\"of dependent type theory for analyzing
event telicity, culminativity, and associated entailments. As far as we know, this is the first
analysis of telicity in the context of dependent type theory. In this paper we focus on our
framework as it applies to English; the first author's dissertation \citep{kovalev2024modeling} also
includes an extended discussion of Russian. In particular, our analysis is careful to distinguish
the phenomena of telicity and culminativity, which (as discussed in Section~\ref{sec:bg-telicity})
generally coincide in English but not in e.g.\ Russian or Mandarin.

Our analysis is based on the well-observed idea that for a large class of verbs, the telicity of an
event correlates with the boundedness of its undergoer, its most patient-like participant
\citep{1972_Verkuyl_BOOK, 1993_Verkuyl_BOOK, 1979_Dowty_BOOK, tenny1987grammaticalizing,
1994_Tenny_BOOK, 1989_Krifka_CONF, 1992_Krifka, 1998_Krifka, 1991_Jackendoff, 1996_Jackendoff}. Much
of this paper is thus dedicated to developing a thorough ontology of the structure of noun phrases,
and particularly of the internal structure of overtly bounded noun phrases: numerical quantifiers,
measure words, adjectival modification, instances of noun phrases, and mereological sums thereof. We
believe this part of our framework is of independent interest.

Following \cite{2017_Luo_CONF}, we consider events as a dependent type family indexed by actors and
undergoers. Unlike previous work in dependent type theory, we regard events as giving rise to their
collections of occurrences, allowing us to easily model adverbial modification. Because the type of
events is indexed by the type of undergoers, and the type of undergoers is further indexed by
boundedness, we are able to detect telicity purely through type dependency. We then associate to
each telic event a resulting state, and define culminativity as the property that a telic event
obtains its resulting state if it occurs.


In addition to modeling telicity and culminativity, we aim to build a reasonable ontology of
eventualities---both static and dynamic---and to advance the use of dependent types for event
structure, to which relatively little attention has been paid. In addition to the dependent event
types of \cite{2017_Luo_CONF}, some preliminary work on this has been done by
\citet{2020_Corfield_BOOK} in his sketch of the modeling of Vendler's Aktionsart classes
\citep{1957_Vendler, 1967_Vendler_BOOK}, as we discuss in Section~\ref{sec:Future}.

Finally, although many aspects of our framework are heavily inspired by MTT-semantics
\citep{2020_Chatzikyriakidis_BOOK}, we restrict ourselves to working inside a standard intensional
type theory \citep{MartinLof75itt} rather than the Unifying Theory of Dependent Types
\citep{1994_Luo_BOOK} extended with various complex features such as coercive subtyping. Using a
standard type theory not only simplifies our framework but has also enabled us to formalize all the
rules and examples in this paper in the Agda proof assistant, and provides a clear path toward a
set-theoretic semantics with respect to which we can validate our entailments. In this respect our
work can be seen as an improvement on lexical (or predicate) decomposition approaches, in that it
provides both a representation language and also an associated semantics.

\paragraph*{Outline}

The rest of the paper is structured as follows.
In Section~\ref{sec:Background}, we provide a very quick summary of intensional Martin-L\"of type
theory, our ambient type theory.
In Sections~\ref{sec:Nominal} and \ref{sec:Verbal} we present the nominal and verbal fragments of
our framework respectively, interspersed with examples. Our nominal framework places a particular
emphasis on tracking the boundedness of noun phrases and particularly in the internal structure of
overtly bounded noun phrases, which we believe to be of independent interest. The verbal framework
builds on the nominal framework to define and analyze eventualities of different kinds---including
static eventualities and telic, atelic, culminating, and non-culminating dynamic eventualities---and
entailments between these.
In Section~\ref{sec:Agda}, we discuss and briefly showcase our Agda formalization of the rules and
examples from Sections~\ref{sec:Nominal} and \ref{sec:Verbal}.
Finally, in Section~\ref{sec:Future} we close with a discussion of possible extensions to our
framework and how it compares to prior work.

\paragraph*{Acknowledgements}

This article extends and reformulates parts of the first author's dissertation
\citep{kovalev2024modeling}. The second author's contributions were supported by the Air Force
Office of Scientific Research under award number FA9550-24-1-0350. We are grateful to Thomas Grano
and Larry Moss for helpful discussions with the first author during the preparation of that
dissertation, particularly concerning linguistics. We also thank the anonymous reviewer for helpful
comments and suggestions.

%% file: sections/background.tex
Our framework builds on predicative intensional Martin-L\"of dependent type theory
\citep{MartinLof75itt}, a standard variant of type theory which corresponds roughly to the
intersection of features available in modern type-theoretic proof assistants, such as Agda
\citep{Agda}, Rocq/Coq \citep{Coq2010}, and Lean \citep{DeMouraUllrich21}. Readers unfamiliar with
type theory are encouraged to consult one of the many references on this theory or variations
thereof, such as the books of \citet{NordstromPeterssonSmith90,1994_Luo_BOOK,NederpeltGeuvers14}. In
this section, we quickly recall some basics and fix notation.

Dependent type theory consists of four main judgments. The first, $\Gamma\vdash A\type$, states that
$A$ is a type relative to context $\Gamma$, a list of variables/hypotheses
$x_1:A_1,x_2:A_2,\dots,x_n:A_n$. The second, $\Gamma\vdash a:A$, states that $a$ is a term of type
$A$, again both relative to context $\Gamma$. We can understand the judgment $\Gamma\vdash a:A$ in
one of two ways: either that $a$ is a proof of the proposition $A$ under a list of hypotheses
$\Gamma$, or that $a$ is a $\Gamma$-indexed family of elements of the ($\Gamma$-indexed) collection
$A$. (In particular, when $\Gamma$ is empty, $\cdot\vdash a:A$ can be understood as a proof of the
proposition $A$, or as an element of the set $A$.) Finally, we have judgmental equalities
$\Gamma\vdash A = A'\type$ and $\Gamma\vdash a = a' : A$ which state that the two types (resp.,
terms) $A$ and $A'$ (resp., $a$ and $a'$) are the same. These four judgments are inductively defined
by a collection of natural deduction--style rules which can be formally understood as collectively
specifying the signature of a generalized algebraic theory \citep{Cartmell86}.

In this paper, we rely almost exclusively on three type formers: $\Pi$-types (dependent products),
$\Sigma$-types (dependent sums), and the universe $\Prop$ of homotopy propositions.\footnote{We also
assume one type universe $\Univ$, intensional identity types $\mathbf{Id}_A(a,a')$, and function
extensionality.} $\Pi$-types are the dependent generalization of ordinary function types, and
correspond logically to universal quantification. If $B(x)$ is a family of types indexed by $x:A$,
then the $\Pi$-type $(x:A)\to B(x)$ consists of functions $\lambda x.b$ which send any $x:A$ to an
element $b$ of $B(x)$. We write $f\ a : B(a)$ for the application of any function $f : (x:A)\to
B(x)$ to an argument $a : A$.
\begin{mathpar}
  \inferrule
  {\Gamma,x:A \vdash B(x) \type}
  {\Gamma \vdash (x:A)\to B(x) \type}
  \and
  \inferrule
  {\Gamma,x:A \vdash b : B(x)}
  {\Gamma \vdash \lambda x.b : (x:A)\to B(x)}
  \and
  \inferrule
  {\Gamma \vdash f : (x:A)\to B(x) \\
   \Gamma \vdash a : A}
  {\Gamma \vdash f\ a : B(a)}
\end{mathpar}

On the other hand, $\Sigma$-types are the dependent generalization of ordinary pair types, and
correspond to a form of existential quantification. If $B(x)$ is a family of types indexed by $x:A$,
then the $\Sigma$-type $\sum_{x:A} B(x)$ consists of pairs of a term $a:A$ and a term $b:B(a)$.
Given $p : \sum_{x:A} B(x)$ we write $\fst(p) : A$ and $\snd(p) : B(\fst(p))$ for its first and
second projections.
\begin{mathpar}
  \inferrule
  {\Gamma,x:A \vdash B(x) \type}
  {\Gamma \vdash \textstyle\sum_{x:A} B(x) \type}
  \and
  \inferrule
  {\Gamma \vdash a : A \\
   \Gamma \vdash b : B(a)}
  {\Gamma \vdash (a,b) : \textstyle\sum_{x:A} B(x)}
  \\
  \inferrule
  {\Gamma \vdash p : \textstyle\sum_{x:A} B(x)}
  {\Gamma \vdash \fst(p) : A}
  \and
  \inferrule
  {\Gamma \vdash p : \textstyle\sum_{x:A} B(x)}
  {\Gamma \vdash \snd(p) : B(\fst(p))}
\end{mathpar}

Our treatment of the type of propositions, $\Prop$, requires more discussion. We model adjectives,
adverbs, and static eventualities using families of ``proof-irrelevant'' propositions, or types that
have at most one term; this ensures that, for instance, there is at most one proof that a given
\emph{cat} is \emph{black}. There are many different versions of $\Prop$ in the type-theoretic
literature, such as homotopy propositions \citep[Section 3.3]{HoTT13}, strict propositions
\citep{GilbertCockx+19}, and impredicative propositions \citep{CoquandHuet88}. Our requirements are
very mild: we need a type $\Prop$ whose elements $P:\Prop$ give rise to types $\Prff(P)$ with at
most one element up to the intensional identity type.
\begin{mathpar}
  \inferrule
  { }
  {\Gamma \vdash \Prop \type}
  \and
  \inferrule
  {\Gamma \vdash P : \Prop}
  {\Gamma \vdash \Prff(P) \type}
  \and
  \inferrule
  {\Gamma \vdash P : \Prop \\
   \Gamma \vdash p : \Prff(P) \\
   \Gamma \vdash p' : \Prff(P)}
  {\Gamma \vdash \mathbf{irr}(p,p') : \mathbf{Id}_{\Prff(P)}(p,p')}
\end{mathpar}

In our Agda formalization, we choose $\Prop$ to be the type of homotopy propositions
\[
  \Prop \triangleq \textstyle\sum_{X : \Univ} (x:X)\to (y:X)\to \mathbf{Id}_{X}(x,y)
\]
and assume function extensionality in Section~\ref{sec:culminating} to show that $X\to\Prff(P)$ is a
proposition when $P:\Prop$. However, any other variation of $\Prop$ satisfying these properties is
equally suitable for our purposes. Notably, unlike \citet{2020_Chatzikyriakidis_BOOK} we do not
require any form of impredicativity in our type theory.

In order to model various linguistic phenomena, our framework extends intensional type theory by
various primitive type and term formers and judgmental equalities over these new primitives, such as
the dependent type $\NP^b$ of noun phrases of a given boundedness $b:\Boundedness$. These extensions
are described throughout Sections~\ref{sec:Nominal} and \ref{sec:Verbal}. In those sections we also
illustrate how to use our framework to analyze fragments of English in a series of examples which
temporarily extend the framework by additional constants modeling English words, such as the
unbounded noun phrase \emph{human} represented by the constant $\human : \NP^\U$.

%% file: sections/nominal.tex
Before developing our account of telicity and culminativity in the verbal domain, we must first lay
some groundwork in the nominal domain. In particular, we refine the \emph{common nouns as types}
paradigm of \citet{2012_Luo_CONF, 2012_Luo, 2020_Chatzikyriakidis_BOOK} by using dependent types to
track the boundedness of noun phrases, which we will use in Section~\ref{sec:Verbal} to detect
telicity.

In Section~\ref{sec:UniversesOfNP} we introduce two types corresponding to the collections of
bounded and unbounded noun phrases respectively. Because we are working in the common nouns as types
paradigm, we associate to every element of these types (such as \emph{human}) a collection of its
instances (in this case, the type of all humans). In Section~\ref{sec:ProperNouns}, we define the
type of proper names, which are pairs of a noun phrase and a particular instance of that noun phrase
(such as \emph{John}, a particular human).

In Section~\ref{sec:Subtyping in the nominal domain}, we define a notion of inclusion between noun
phrases, allowing us to express the relation that (e.g.) \emph{every man is a human}. Note that
terms have at most one type in our setting of Martin-L\"of type theory; we do not express inclusion
in terms of coercive subtyping \citep{Luo1997,1999_Luo,2013_Luo} as in MTT-semantics
\citep{2009_Luo_CONF, 2012_Luo, 2020_Chatzikyriakidis_BOOK}, nor can we refer to a subset relation
as one might in set theory.

In our framework, bare nouns in English such as \emph{human}, whether mass or count, correspond to
\emph{unbounded} noun phrases. In Section~\ref{sec:Degrees and Units}, we discuss the origin of
\emph{bounded} noun phrases as those which express some delimited quantity, such as \emph{three
apples}. These quantities are measured in terms of degrees and units, following \cite{1987_Loenning} and
\cite{2017_Champollion_BOOK}. We focus primarily on the case in which
bounded noun phrases have an overt determiner, as in English; however, our framework can also account
for article-less languages such as Russian (Remark~\ref{rem:several}), where bare nouns can be used in a bounded sense. In this section we also account for the distinction between mass and
count nouns and model a way to obtain composite individuals from primitive individuals, laying the
groundwork for future potential analyses of distributivity phenomena.


We believe that Section~\ref{sec:Degrees and Units} is of independent interest beyond its role in
our framework of generating bounded noun phrases, because the internal structure of overtly bounded
noun phrases has not received much attention in the context of dependent types. As far as we know,
our presentation is the first more or less complete treatment of the internal structure of noun
phrases in the setting with dependent types.


Finally, in Section~\ref{sec:Adjectival modification}, we discuss adjectives, paying particular
attention to the boundedness of adjectivally-modified noun phrases. We limit our study to
intersective adjectives and their interaction with the rest of our framework, namely boundedness,
the subtyping relation of Section~\ref{sec:Subtyping in the nominal domain}, the property of being
count, and composite individuals.

\subsection{Universes in the nominal domain}\label{sec:UniversesOfNP}

As discussed in Section~\ref{sec:MTTsemantics}, standard MTT-semantics
\citep{2020_Chatzikyriakidis_BOOK} includes a universe $\CN$ of common nouns (possibly modified by adjectives). Our counterpart is the universe $\NP$ of noun phrases, but we crucially 
divide $\NP$ into two subuniverses: the universe $\NP^{\B}$ of \emph{bounded noun phrases} and the
universe $\NP^{\U}$ of \emph{unbounded noun phrases}.

\begin{remark}\label{rem:NPterminology}
Note that we use the term ``noun phrase'' not in the same sense as it is used in generative syntax.
Specifically, we do not consider proper names as noun phrases; instead, we assign them the separate
type of $\Entity$, which we introduce in Section~\ref{sec:ProperNouns}.
\end{remark}


First, we introduce the type $\Boundedness$ of ``kinds of boundedness,'' with two elements: $\B$ (for
bounded) and $\U$ (for unbounded). This is essentially the type of booleans with a different name,
although we will not consider any case analysis on $\Boundedness$ until Example~\ref{ex:type of
pop}.
\begin{mathpar}
  \inferrule
  { }
  {\Gamma\vdash \Boundedness \type}
  \and
  \inferrule
  { }
  {\Gamma\vdash\B: \Boundedness}
  \and
  \inferrule
  { }
  {\Gamma\vdash\U: \Boundedness}
\end{mathpar}

For any $b:\Boundedness$ we introduce a type $\NP^b$ of ``NPs with boundedness $b$,'' or $b$-noun
phrases:
\[
  \inferrule
  {\Gamma \vdash b:\Boundedness}
  {\Gamma\vdash \NP^b \type}
\]
A ``bounded noun phrase'' is a term of type $\NP^\B$ and an ``unbounded noun phrase'' is a term of
type $\NP^\U$. We associate to every $b$-noun phrase $\var{np}:\NP^b$ a type of ``instances of
$\var{np}$,'' written $\El_\NP^b(\var{np})$, for the collection of ``elements of'' $\var{np}$.
\[
  \inferrule
  {\Gamma\vdash b:\Boundedness \\
   \Gamma\vdash \var{np}:\NP^b}
  {\Gamma\vdash \El_{\NP}^b(\var{np}) \type}
\]

\begin{notation}\label{notation:fonts}
We use \textbf{bold} for term and type constructors that are part of our framework, and
\textsf{sans-serif} for term constants introduced to model a fragment of language using our
framework. We use \textit{italics} to refer to arbitrary terms (as in the premises of inference
rules).
\end{notation}

The type $\NP$ of all noun phrases is defined as the indexed sum ($\Sigma$-type) over all
$b:\Boundedness$ of $\NP^b$, so that terms of type $\NP$ are pairs $(b,\var{np})$ of a boundedness
$b:\Boundedness$ along with some $b$-noun phrase $\var{np}:\NP^b$. The type $\El_\NP(\var{n})$ of
instances of a noun phrase $n:\NP$ is definable in terms of the primitive type families $\El_\NP^b$
stipulated above.
\begin{mathpar}
  \inferrule
  { }
  {\Gamma\vdash \NP \triangleq \textstyle\sum_{b:\Boundedness} \NP^b \type}
  \and
  \inferrule
  {\Gamma\vdash n:\NP}
  {\Gamma\vdash \El_{\NP}(n) \triangleq \El_{\NP}^{\fst(n)}(\snd(n)) \type}
\end{mathpar}

Likewise, without extending our framework, we can regard any $b$-noun phrase as a noun phrase. We
say that terms of type $\NP^b$ can be ``lifted'' to type $\NP$.
\[
  \inferrule
  {\Gamma\vdash b:\Boundedness \\
   \Gamma\vdash \var{np}:\NP^b}
  {\Gamma\vdash \Lift_{\NP}^b(\var{np}) \triangleq (b,\var{np}) : \NP}
\]
Note that $\Lift_\NP^b$ is well-typed by the definition of $\NP$ as a $\Sigma$-type. In addition, by
expanding definitions it is easy to see that $\El_{\NP}(\Lift_{\NP}^b(\var{np})) =
\El_{\NP}^b(\var{np})$ for any $b:\Boundedness$ and $\var{np}:\NP^b$.

Finally, we require that $\NP^b$ be closed under $\Sigma$-types of predicates over $\El_\NP^b$. That
is, for any $b$-noun phrase $\var{np}:\NP^b$ and any family of propositions $P$ indexed by
$\El_\NP^b(\var{np})$, we require that ``the subset of \var{np}s satisfying $P$'' is again a
$b$-noun phrase. This is a generalization of the condition that noun phrases modified by adjectives
are again noun phrases (in particular, it is not limited to adjectival modification). Note also
that this operation preserves boundedness.
\begin{mathpar}
  \inferrule
  {\Gamma\vdash b:\Boundedness \\
   \Gamma\vdash \var{np}:\NP^b \\
   \Gamma,p:\El_{\NP}^b(\var{np})\vdash P : \Prop}
  {\Gamma\vdash \sumNP_{p:\var{np}} P : \NP^b}
  \and
  \inferrule
  {\Gamma\vdash b:\Boundedness \\
   \Gamma\vdash \var{np}:\NP^b \\
   \Gamma,p:\El_{\NP}^b(\var{np})\vdash P : \Prop}
  {\Gamma\vdash \El_{\NP}^b\left(\sumNP_{p:\var{np}} P\right) =
   \textstyle\sum_{p:\El_{\NP}^b(\var{np})} \Prff(P) \type}
\end{mathpar}


When modeling a particular fragment of English, we will populate $\NP^\U$ with bare nouns as well
as bare nouns modified by intersective adjectives (Section~\ref{sec:Adjectival modification}). Thus,
$\NP^\U$ plays the same role as $\CN$ in \cite{2020_Chatzikyriakidis_BOOK}.

\begin{example}\label{ex:john is a term of type El human}
Suppose we are interested in analyzing a fragment of English containing the word \emph{human}. We
extend our framework with a term
\[
  \inferrule
  { }
  {\Gamma \vdash \human:\NP^{\U}}
\]
corresponding to the unbounded noun phrase \emph{human}. The type $\El_{\NP}^\U(\human)$ is then the
type of all humans. To further include the word \emph{John}---a particular instance of a human---in
our fragment, we would extend our framework with a second term
\[
  \inferrule
  { }
  {\Gamma\vdash \john : \El_{\NP}^{\U}(\human)}
\]
\end{example}

\subsection{Entities}\label{sec:ProperNouns}

With respect to telicity, proper names such as John
(Example~\ref{ex:john is a term of type El human}) behave in the same way as bounded noun phrases. However,
we do not assert that $\john$ has type $\NP^\B$. For one, our framework assigns $\john$ the type
$\El_{\NP}^{\U}(\human)$, and it cannot have two different types at the same time. Nor do we wish to
assert a lift $\iota : \El_{\NP}^\U(\human) \to \NP^\B$ stating that every instance of $\human$s is
a $\NP^\B$, because this forces us to consider the type $\El_\NP^\B(\iota(\john))$ of ``instances of
$\john$s,'' which is semantically questionable.

Instead, we introduce a new type of proper names, or \emph{entities}, whose elements are pairs of a
noun phrase $n:\NP$ of any boundedness along with an instance $p:\El_\NP(n)$ of $n$. In fact, the
type of entities is once again already definable as a $\Sigma$-type:
\[
  \inferrule
  { }
  {\Gamma \vdash  \Entity \triangleq \textstyle\sum_{n:\NP} \El_{\NP}(n) \type}
\]
Expanding the definition of $\NP$ from Section~\ref{sec:UniversesOfNP}, terms of type $\Entity$ are
tuples $((b,\var{np}),p)$ where $b : \Boundedness$, $\var{np} : \NP^b$, and $p :
\El_{\NP}^b(\var{np})$. For example, $((\U,\human),\john):\Entity$ is the entity corresponding to
the human named John.

Entities will play an important role in our treatment of adjectival modification
(Section~\ref{sec:Adjectival modification}) and in our formulation of dependent event types in
Section~\ref{sec:Verbal}.

\subsection{Subtyping in the nominal domain}\label{sec:Subtyping in the nominal domain}

Next, we need a way of asserting (e.g.) that every man can be regarded as a human. We define a
family of types $\isA(\var{np},\var{np}')$ capturing the proposition that \emph{Every $\var{np}$ is
an $\var{np}'$} for $\var{np},\var{np}'$ noun phrases of arbitrary (and in particular, potentially
different) boundedness:
\[
  \inferrule
  {\Gamma\vdash b:\Boundedness \\
   \Gamma\vdash \var{np}: \NP^b \\
   \Gamma\vdash b':\Boundedness \\
   \Gamma\vdash \var{np}': \NP^{b'}}
  {\Gamma\vdash \isA(\var{np},\var{np}') \type}
\]

Proofs of $\isA(\var{np},\var{np}')$ induce inclusions
$\El_\NP^b(\var{np})\to\El_\NP^{b'}(\var{np}')$ from the type of instances of $\var{np}$s to the
type of instances of $\var{np}'$s.\footnote{We represent $\isA(\var{np},\var{np}')$ as a type rather
than a term of type $\Prop$ because some type theories with $\Prop$ restrict the ability to define
functions from a proposition into a non-proposition (in this case, $\El_{\NP}^b(\var{np})\to
\El_{\NP}^{b'}(\var{np}')$).} We write $\El_{\isA}$ for the map from such proofs to such
inclusions:\footnote{For simplicity we only assert that $\El_\isA(\var{prf})$ is a \emph{function},
but one could moreover assert that it is an \emph{injective function}.}
\[
  \inferrule
  {\Gamma\vdash b:\Boundedness \\
   \Gamma\vdash \var{np}: \NP^b \\
   \Gamma\vdash b':\Boundedness \\
   \Gamma\vdash \var{np}': \NP^{b'} \\
   \Gamma\vdash \var{prf} : \isA(\var{np},\var{np}')}
  {\Gamma\vdash \El_{\isA}(\var{prf}) : \El_{\NP}^b(\var{np})\to \El_{\NP}^{b'}(\var{np}')}
\]

\begin{example}
Suppose we are interested in modeling a fragment of English with \emph{human}, \emph{woman},
\emph{Ann} (as an instance of \emph{woman}), and \emph{talk}, the latter being a predicate over
(instances of) \emph{humans}.
\begin{mathpar}
  \inferrule
  { }
  {\Gamma\vdash \human : \NP^\U}
  \and
  \inferrule
  { }
  {\Gamma\vdash \woman : \NP^\U}
  \\
  \inferrule
  { }
  {\Gamma\vdash \ann : \El_{\NP}^\U(\woman)}
  \and
  \inferrule
  { }
  {\Gamma\vdash \talk:\El^\U(\human)\to \Prop}
\end{mathpar}

To model the fact that every woman is a human, we further add
\[
  \inferrule
  { }
  {\Gamma\vdash \textsf{womanIsHuman} : \isA(\woman, \human)}
\]
giving us an inclusion $\El_{\isA}(\textsf{womanIsHuman}):\El_\NP^\U(\woman)\to \El_\NP^\U(\human)$
which allows us to regard $\ann$ as a $\human$:
\[
  \textsf{annQuaHuman} \triangleq \El_{\isA}(\textsf{womanIsHuman})\ \ann : \El_\NP^\U(\human)
\]

We may then state the proposition \emph{Ann talks} as $\talk(\textsf{annQuaHuman})$. Note that
$\talk(\ann)$ is not well-formed because $\talk$ is a predicate over humans and not women.
\end{example}

Finally, we require that $\isA$ be reflexive and transitive.\footnote{One could moreover assert that
$\El_\isA(\isArefl(\var{np}))$ is the identity function on $\El_\NP^b(\var{np})$ and that
$\El_\isA(\isAtrans(\var{prf},\var{prf}\;'))$ is $\El_\isA(\var{prf}\;') \circ \El_\isA(\var{prf})$,
but we omit these for simplicity.}
\begin{mathpar}
  \inferrule
  {\Gamma\vdash b : \Boundedness \\
   \Gamma\vdash \var{np} :\NP^b}
  {\Gamma\vdash \isArefl(\var{np}) : \isA(\var{np},\var{np})}
  \and
  \inferrule
  {\Gamma\vdash b,b',b'':\Boundedness \\
   \Gamma\vdash \var{np} : \NP^b \\
   \Gamma\vdash \var{np}' : \NP^{b'} \\
   \Gamma\vdash \var{np}'' : \NP^{b''} \\\\
   \Gamma\vdash \var{prf} : \isA(\var{np},\var{np}') \\
   \Gamma\vdash \var{prf}\;' : \isA(\var{np}',\var{np}'')}
  {\Gamma\vdash \isAtrans(\var{prf},\var{prf}\;'): \isA(\var{np},\var{np}'')}
\end{mathpar}

\subsection{Internal structure of overtly bounded noun phrases}\label{sec:Degrees and Units}

In Section~\ref{sec:UniversesOfNP}, we said that when modeling a particular fragment of English, the
type $\NP^\U$ will be populated by bare noun phrases. In contrast, $\NP^\B$ will be populated by
terms corresponding to noun phrases which express some delimited quantity,\footnote{As mentioned in
Section~\ref{sec:Relationship between telicity and undergoers}, in languages which lack articles,
bare noun phrases may have definite readings (unlike in English) and hence they may have bounded
readings in addition to unbounded readings.  We will discuss the case of article-less languages in
Section~\ref{sec:ambiguousNPs}.} including noun phrases such as \emph{3 apples, 3 kilograms of
apples, a couple of apples, the apples, an apple, few apples}, etc. Before introducing our
representation of such noun phrases, we must discuss degrees and units.

Degrees have been used in the linguistic literature to model gradable adjectives and nouns as well as measure and comparative constructions. We will be only concerned with measure constructions. Our treatment of degrees and units will be close to \cite{2017_Champollion_BOOK} and \cite{1987_Loenning}. However,
we will take degrees to be fully primitive, as in \cite{2020_Chatzikyriakidis_BOOK}; we will not
think of degrees as things in the range of measure functions, as it is done in
\cite{2017_Champollion_BOOK}.  Specifically, we declare a type $\Degree$ of degrees, which will
contain primitive terms such as $\height, \weight, \width, \vvolume, \quantity$, etc.  ($\quantity$
in particular will play an important role momentarily.) This is the type that is needed to
distinguish between, for example, \emph{3 apples} and \emph{3 kilograms of apples}, where in the
former case the relevant degree is quantity, whereas in the latter, it is weight.
\begin{mathpar}
  \inferrule
  { }
  {\Gamma\vdash \Degree \type}
  \and
  \inferrule
  { }
  {\Gamma\vdash \quantity :\Degree}
\end{mathpar}

Next, for any $d:\Degree$ we have a type $\Units(d)$ of units for $d$. This allows us to distinguish
between, for example, \emph{50 kilograms of apples} and \emph{50 grams of apples}. To ensure a
uniform treatment, we adopt the concept of \emph{natural units}, written $\natunits$, as proposed in
\cite{1989_Krifka_CONF} when dealing with noun phrases like \emph{five apples}. Examples of units
include $\meter : \Units(\height)$, $\kilogram : \Units(\weight)$, $\liter : \Units(\vvolume)$,
$\natunits : \Units(\quantity)$, etc. ($\natunits$ will also play an important role in the framework
shortly.)
\begin{mathpar}
  \inferrule
  {\Gamma\vdash d:\Degree}
  {\Gamma\vdash \Units(d) \type}
  \and
  \inferrule
  { }
  {\Gamma\vdash \natunits : \Units(\quantity)}
\end{mathpar}

\begin{remark}
Note that the dependency of $\Units$ on $\Degree$s allows us to express that each unit is only
compatible with certain degrees; for example, one cannot measure volume in meters, so $\meter$ is
not a term of type $\Units(\vvolume)$. In contrast, for us $\Degree$s do not depend on the noun
being measured, so we permit phrases such as \emph{three kilograms of books}. One might imagine that
$\Degree$ (and thus $\Units$) should depend on $\NP^\U$, so that $\weight$ is not a valid degree for
$\book : \NP^\U$, but we have opted not to model this. Note that---whereas volume cannot be measured in meters under any circumstances---the phrase
\emph{three kilograms of books} is perfectly meaningful, describing a collection of books whose total
weight is three kilograms.

Our notion of units can also account for less conventional units, such as \emph{glass} in \emph{five glasses of water}, as well as for the so-called nominal classifiers, which are rare in English (one example from \cite{1989_Krifka_CONF} being \emph{head} in \emph{five head of cattle}) but are abundant in languages like Chinese. However, if one wishes to consider classifiers more seriously, one may need to introduce the above-mentioned dependence of $\Units$ on $\NP^{\U}$, since numeral classifiers usually depend on the noun they classify.
\end{remark}

Finally, for any unbounded noun phrase $\var{np}:\NP^\U$ and any $d:\Degree$, $u:\Units(d)$, and
natural number $m:\Nat$, we introduce a \emph{bounded} noun phrase
$\AmountOf(\var{np},d,u,m):\NP^\B$ for the corresponding amount of $\var{np}$. 
%
\[
  \inferrule
  {\Gamma\vdash \var{np}:\NP^\U \\
   \Gamma\vdash d:\Degree \\
   \Gamma\vdash u: \Units(d) \\
   \Gamma\vdash m:\Nat}
  {\Gamma\vdash \AmountOf(\var{np},d,u,m):\NP^\B}
\]

\begin{example}\label{ex:degrees}
To model a fragment of English containing the noun phrases \emph{3 humans} and \emph{3 kilograms of
apples}, we additionally assert the following rules:
\begin{mathpar}
  \inferrule
  { }
  {\Gamma\vdash \apple : \NP^\U}
  \and
  \inferrule
  { }
  {\Gamma\vdash \human : \NP^\U}
  \\
  \inferrule
  { }
  {\Gamma\vdash \weight : \Degree}
  \and
  \inferrule
  { }
  {\Gamma\vdash \kilogram : \Units(\weight)}
\end{mathpar}
Using the framework rules described above, we obtain terms
\begin{gather*}
\textsf{threeHumans} \triangleq \AmountOf(\human,\quantity,\natunits, 3):\NP^\B
\\
\textsf{threeKgApples} \triangleq \AmountOf(\apple,\weight,\kilogram, 3):\NP^\B
\end{gather*}
respectively encoding the English phrases \emph{3 humans} and \emph{3 kilograms of apples}. Terms of
type $\El_{\NP}^\B(\textsf{threeHumans})$ are particular 3 humans, such as John, Mary, and Susan;
likewise, terms of type $\El_{\NP}^\B(\textsf{threeKgApples})$ are particular 3 kilograms of apples
(although, unlike humans, one typically does not give names to particular instantiations of 3
kilograms of apples).

We do not distinguish between, for example, \emph{an apple, the apple}, and \emph{one apple}; the
first two noun phrases are modeled as the third one.
\end{example}

\subsubsection{Count nouns}

In light of Example~\ref{ex:john is a term of type El human}, we now have two ways of representing a
particular human, such as John: as terms of type $\El_{\NP}^\U(\human)$ and as terms of type
$\El_{\NP}^\B(\AmountOf(\human,\quantity,\natunits, 1))$. The same applies to other count nouns and
count nouns modified by adjectives (collectively, \emph{count noun phrases}), but not to mass nouns.
In fact, we will never consider instances of mass nouns (e.g., terms of type $\El_\NP^\U(\water)$)
because (unlike for \emph{human}) it is not clear how one should think of an instance of
\emph{water}.%
\footnote{One can have a reading of \emph{water} which is forced into the count reading, as in
\emph{Could I just get one water, please}. In this case we represent \emph{one water} as a term of
type $\El_{\NP}^\B(\AmountOf(\water, \quantity, \natunits, 1))$. Additionally, the so-called object mass
nouns such as \emph{furniture} admit grammatically accessible natural units, as evidenced by their
compatibility with stubbornly distributive predicates (e.g., \emph{round furniture}) and by
cardinality readings in \emph{more}-constructions
\citep{1989_Krifka_CONF,schwarzschild2011stubborn,suttonfilip2018subkind}. Conversely, certain count
nouns such as \emph{fence}, \emph{road}, and \emph{sequence} lack clear natural units; their
individuation is context dependent
\citep{1989_Krifka_CONF,rothstein2010counting,rothstein2017semantics}.  A full treatment of this
distinction lies beyond the scope of the present work.}

To enforce that $\El_{\NP}^\U(\human)$ and $\El_{\NP}^\B(\AmountOf(\human,\quantity,\natunits, 1))$
are the same, we first define a predicate expressing that an unbounded noun phrase is count:
\[
  \inferrule
  {\Gamma\vdash \var{np}:\NP^\U}
  {\Gamma\vdash \isCount(\var{np}) :\Prop}
\]
Then, for any count noun phrase $\var{np}$, we say that a $\var{np}$ is a
$\AmountOf(\var{np},\quantity, \natunits, 1)$ and vice versa.
\begin{mathpar}
  \inferrule
  {\Gamma\vdash \var{np}:\NP^\U \\
   \Gamma\vdash \var{prf}:\Prff(\isCount(\var{np}))}
  {\Gamma\vdash \NPIsOneNP(\var{np},\var{prf}):\isA(\var{np}, \AmountOf(\var{np},\quantity,\natunits,1))}
  \and
  \inferrule
  {\Gamma\vdash \var{np}:\NP^\U \\
   \Gamma\vdash \var{prf}:\Prff(\isCount(\var{np}))}
  {\Gamma\vdash \OneNPIsNP(\var{np},\var{prf}):\isA(\AmountOf(\var{np},\quantity,\natunits,1),\var{np})}
\end{mathpar}

Count noun phrases are also closed under $\Sigma$-types: if $\var{np}$ is a count noun phrase, then
``the subset of \var{np}s satisfying $P$'' (for any predicate $P$) is also a count noun phrase:
\[
  \inferrule
  {\Gamma\vdash \var{np}:\NP^\U \\
   \Gamma\vdash \var{prf}:\isCount(\var{np}) \\
   \Gamma\vdash P : \El_\NP^\U(\var{np})\to \Prop}
  {\Gamma\vdash \SumIsCount(\var{np},\var{prf},P) :
   \Prff\left(\isCount\left(\sumNP_{p:\var{np}} P(p)\right)\right)}
\]

In Example~\ref{ex:degrees} we said that terms of type
$\El_{\NP}^\B(\AmountOf(\human,\quantity,\natunits, 3))$ correspond to a particular three humans,
but we have not yet introduced any mechanism for deriving such a term from three particular
individual humans. To rectify this, we introduce:
\[
  \inferrule{
    \Gamma \vdash \var{np}:\NP^\U \\
    \Gamma \vdash d:\Degree \\
    \Gamma \vdash u:\Units(d) \\
    \Gamma \vdash m, n : \Nat \\
    \Gamma \vdash p:\El_{\NP}^\B(\AmountOf(\var{np},d,u,m)) \\
    \Gamma \vdash q:\El_{\NP}^\B(\AmountOf(\var{np},d,u,n))
  }{
    \Gamma \vdash p \oplus q : \El_{\NP}^\B(\AmountOf(\var{np},d,u,m+n))
  }
\]

\begin{example}\label{ex:john+mary}
We construct a term of type $\El_{\NP}^\B(\AmountOf(\human,\quantity,\natunits, 2))$ out of two
terms of type $\El_{\NP}^\U(\human)$, using $\oplus$ and $\NPIsOneNP$. Suppose we have:
\begin{mathpar}
  \inferrule{ }{\Gamma\vdash \human : \NP^\U}
  \and
  \inferrule{ }{\Gamma\vdash \prf: \Prff(\isCount(\human))}
  \\
  \inferrule{ }{\Gamma\vdash \john: \El_{\NP}^\U(\human)}
  \and
  \inferrule{ }{\Gamma\vdash \mary: \El_{\NP}^\U(\human)}
\end{mathpar}

Note that we cannot directly apply $\oplus$ to $\john$ and $\mary$, because they do not have the
appropriate type for arguments to $\oplus$. However, note that $\NPIsOneNP$ (and
$\prf:\Prff(\isCount(\human))$) induces a function from $\El_{\NP}^\U(\human)$ to the appropriate
type. Writing $\textsf{oneHuman} \triangleq \AmountOf(\human, \quantity, \natunits, 1)$,
\[
\inferrule*
  {\inferrule*
     {\inferrule*{ }{\Gamma\vdash \human : \NP^\U} \\
      \inferrule*{ }{\Gamma\vdash \prf: \Prff(\isCount(\human))}}
     {\Gamma\vdash \NPIsOneNP(\human, \prf): \isA(\human, \textsf{oneHuman})}}
  {\Gamma\vdash f \triangleq \El_\isA(\NPIsOneNP(\human, \prf)) :
   \El_{\NP}^\U(\human) \to \El_\NP^\B(\textsf{oneHuman})}
\]

Thus $f(\john),f(\mary) : \El_{\NP}^\B(\textsf{oneHuman})$, so we have
\[
  f(\john)\oplus f(\mary) : \El_{\NP}^\B(\AmountOf(\human, \quantity, \natunits, 2))
\]
representing \emph{John and Mary}.\footnote{%
Unlike mereological approaches, our framework has $f(\john)\oplus f(\john) :
\El_{\NP}^\B(\AmountOf(\human, \quantity, \natunits, 2))$ by the same reasoning, which is
undesirable. This oversimplification does not affect our analysis of telicity. As future work, one
could enhance the last ($\Nat$) argument of $\AmountOf$ to explicitly track the identity of the
involved individuals and restrict the type of $\oplus$ accordingly.}
\end{example}

The above argument generalizes to all count nouns. As for mass nouns, because we do not consider
instances of mass nouns (such as \emph{water}) we cannot sum such instances; however, we may
consider instances of \emph{three liters of water}, and we can sum instances of \emph{three liters
of water} and \emph{four liters of water} to obtain an instance of \emph{seven liters of water}.

\subsubsection{Ambiguous noun phrases}\label{sec:ambiguousNPs}

So far, we have populated $\NP^\B$ with noun phrases expressing some delimited quantity of some
$\NP^\U$, where elements of $\NP^\U$ are bare noun phrases. However, as noted in
Section~\ref{sec:Relationship between telicity and undergoers}, bare noun phrases in article-less
languages may have definite readings, and thus may have bounded readings (in addition to unbounded
readings) despite not explicitly expressing a delimited quantity. For example, the bounded reading
of the Russian counterpart of \emph{apples}, can mean either \emph{the apples}, or something like
\emph{n apples for some n}, as discussed in \cite{kovalev2024modeling}. Modeling the former reading
would require cross-sentential relationships, which is out of scope of this work, but it is
straightforward to account for the latter reading. Namely, we introduce a new form of bounded noun
phrase:
\[
\inferrule
  {\Gamma\vdash \var{np}:\NP^\U \\
   \Gamma\vdash d:\Degree \\
   \Gamma\vdash u:\Units(d)}
  {\Gamma\vdash \several(\var{np},d,u) : \NP^\B}
\]
For example, $\several(\apple, \quantity, \natunits) : \NP^\B$ represents the English phrase
\emph{several apples}.

To capture that \emph{several apples} means \emph{n apples for some n}, we add an equation
specifying that the type of instances of the former is the type of instances of the latter, where
\emph{for some $n$} is represented as a $\Sigma$-type:
\[
\inferrule
  {\Gamma\vdash \var{np}:\NP^\U \\
   \Gamma \vdash d:\Degree \\
   \Gamma\vdash u:\Units(d)}
  {\Gamma\vdash \El_{\NP}^\B(\several(\var{np},d,u)) =
   \textstyle\sum_{n:\Nat} \El_{\NP}^\B(\AmountOf(\var{np}, d, u, n)) \type}
\]

\begin{remark}\label{rem:several}
While modeling definiteness is out of scope of this paper, we can use  $\several$ to approximate the
meaning of definite noun phrases in English. For example, we can model \emph{the water} (as well as
the bounded reading of the bare noun phrase corresponding to \emph{water} in article-less languages)
as \emph{n liters of water for some n}. This is a heavy oversimplification, but if one is only
interested in telicity, it is an acceptable oversimplification because both \emph{the water} and
\emph{n liters of water for some n} yield events with the same telicity status. However, this raises
an important question: why did we pick volume and \emph{liters} as the contextually relevant degree
and units? A similar question---when dealing with the bare noun phrase corresponding to \emph{water}
in article-less languages, how do we know whether to use the bounded or unbounded representation of
that word?

Our answer is that, in formal semantics, all ambiguities must be resolved before modeling a fragment
of natural language. So before even modeling a sentence like \emph{John drank water} in an
article-less language, we must decide (e.g., based on the prior discourse) whether the undergoer has
a bounded or unbounded reading, and in the latter case, what are the relevant degree and units. If
we disambiguate \emph{water} as having an unbounded reading, then we represent it as $\water :
\NP^\U$; if we instead disambiguate it as bounded, with volume and liters as its degree and units
respectively, then we represent it as $\several(\water, \vvolume, \liter) : \NP^\B$.  \end{remark}

\subsection{Adjectival modification}\label{sec:Adjectival modification}

In this section, we refine the MTT-semantics approach to modification by intersective adjectives
\citep{Chatzikyriakidis2013, 2017_Chatzikyriakidis, 2020_Chatzikyriakidis_BOOK}, paying particular
attention to the boundedness of adjectivally modified noun phrases. We will not consider other kinds
of adjectives, leaving them for future work.\footnote{Some proposals on how subsective, privative,
and non-committal adjectives should be modeled in MTT-semantics are given in
\citet{Chatzikyriakidis2013, 2017_Chatzikyriakidis, 2020_Chatzikyriakidis_BOOK}.}

We introduce a type $\IntAdj$ of intersective adjectives, and assert that every $\var{adj}:\IntAdj$
gives rise to a predicate over entities, i.e., a function $\Entity\to\Prop$:
\begin{mathpar}
  \inferrule
  { }
  {\Gamma \vdash \IntAdj \type}
  \and
  \inferrule
  {\Gamma\vdash \var{adj}:\IntAdj}
  {\Gamma \vdash \El_{\IA}(\var{adj}) : \Entity\to\Prop}
\end{mathpar}

Intersective adjectival restriction preserves boundedness. Unbounded noun phrases such as bare plurals and mass nouns in English (e.g., \emph{cats}) remain unbounded under intersective restriction (e.g., \emph{black cats}). Likewise, adding an intersective adjective does not affect whether a quantity expression yields a bounded noun phrase: \emph{two cats} and \emph{two black cats} are both bounded. This is because \emph{cats} and \emph{black cats} both yield atelic events, whereas \emph{two cats} and \emph{two black cats} both yield telic events.

Recall from Section~\ref{sec:UniversesOfNP} that for any predicate $P$ over instances of a $b$-noun
phrase $\var{np}$, the subset of $\var{np}$s satisfying $P$, $\sumNP_{p:\var{np}} P$, is again a
$b$-noun phrase. In particular, for any $b:\Boundedness$, $\var{np}:\NP^b$, and intersective
adjective $\var{adj} : \IntAdj$, we have another $b$-noun phrase
\[
  \var{np}' \triangleq \sumNP_{p:\var{np}} \left(\El_{\IA}(\var{adj})\ ((b,\var{np}),p)\right)
  : \NP^b
\]
whose instances are pairs of an instance $p$ of $\var{np}$ along with a proof that $p$, regarded as
an $\Entity$, satisfies the predicate $\El_\IA(\var{adj})$:
\[
  \El_\NP^b(\var{np}') =
  \textstyle\sum_{p:\El_{\NP}^b(\var{np})} \Prff(\El_\IA(\var{adj})\ ((b,\var{np}),p)) \type
\]

\begin{example}\label{ex:black cats formal}
Let's illustrate how to represent the (unbounded) noun phrase \emph{black cats}. First we declare:
\begin{mathpar}
  \inferrule
  { }
  {\Gamma\vdash \cat : \NP^\U}
  \and
  \inferrule
  { }
  {\Gamma\vdash \blk : \IntAdj}
\end{mathpar}

Expanding the definitions of $\El_{\IA}$ and $\Entity$, $\El_{\IA}(\blk)$ is a predicate over pairs
of an $\NP$ and an instance of that $\NP$:
\[
  \El_{\IA}(\blk) : \left(\textstyle\sum_{n:\NP} \El_{\NP}(n)\right) \to \Prop
\]
On the other hand, expanding the definitions of $\NP$ and $\El_{\NP}$:
\[
  \lambda p.((\U,\cat),p) :
  (p : \El^\U_\NP(\cat)) \to \left(\textstyle\sum_{n:\NP} \El_{\NP}(n)\right)
\]
By composition, $\lambda p.\El_{\IA}(\blk)((\U,\cat),p)$ is therefore a predicate over cats:
\[
  \lambda p.\El_{\IA}(\blk)\ ((\U,\cat),p) : \El^\U_\NP(\cat) \to \Prop
\]

We thus model \emph{black cats} as the noun phrase---automatically forced by our framework to be
unbounded---corresponding to the subset of \emph{cats} that are \emph{black}:
\[
  \blackCat \triangleq \sumNP_{p:\cat} \left(\El_{\IA}(\blk)\ ((\U,\cat),p)\right)
  : \NP^\U
\]

An instance of $\blackCat$ is a term of type $\El_\NP^\U(\blackCat)$, which computes
to:
\[
  \El_\NP^\U(\blackCat) =
  \textstyle\sum_{p:\El_{\NP}^\U(\cat)} \Prff(\El_\IA(\blk)\ ((\U,\cat),p)) \type
\]
Terms of this type are pairs $(p,\var{prf})$ of a cat $p : \El_{\NP}^\U(\cat)$ and a witness
$\var{prf} : \Prff(\El_{\IA}(\blk)\ ((\U,\cat),p))$ that $p$ is black. For example, if we further
declare
\begin{mathpar}
  \inferrule
  { }
  {\Gamma\vdash \tom:\El_{\NP}^\U(\cat)}
  \and
  \inferrule
  { }
  {\Gamma\vdash \tomIsBlack : \Prff(\El_{\IA}(\blk)\ ((\U,\cat),\tom))}
\end{mathpar}
then $(\tom,\tomIsBlack) : \El_\NP^\U(\blackCat)$ is an instance of a black cat.
\end{example}

There are several properties of intersective adjectives that we want to ensure are modeled
accurately by our framework. We illustrate these properties in the context of
Example~\ref{ex:black cats formal}:
\begin{enumerate}[label=(\roman*)]
\item a \emph{black cat} is \emph{black};
\item two \emph{black cats} are the same just in case they are the same as \emph{cats};
\item a \emph{black cat} is a \emph{cat};
\item if every \emph{cat} is an \emph{animal}, then a \emph{black cat} is a \emph{black animal};
\item two \emph{black cats} are \emph{two black cats}; and
\item two \emph{black cats} are a \emph{black} instance of \emph{two cats}.
\end{enumerate}

Property (i) holds by definition given the way we represent \emph{black cats} as a pair of a cat and
a proof that the corresponding entity is black. Property (ii) follows from the fact that
$\El_\IA(\var{adj})$ is $\Prop$-valued: in type theory, two pairs are equal if and only if each of
their projections are equal, and any two terms of type $\Prff(\El_\IA(\blk)\ e)$ are equal (where
$e:\Entity$).

For Property (iii), we assert that any instance of a noun phrase
modified by an intersective adjective ``$\isA$'' instance of the underlying noun phrase, in the
sense of Section~\ref{sec:Subtyping in the nominal domain}:
\[
  \inferrule
  {\Gamma\vdash b:\Boundedness \\
   \Gamma\vdash \var{np}:\NP^b \\
   \Gamma\vdash \var{adj}:\IntAdj}
  {\Gamma\vdash \IANPIsNP(\var{np},\var{adj}) :
   \isA\left(\sumNP_{p:\var{np}} \El_{\IA}(\var{adj})\ ((b, \var{np}),p), \var{np}\right)}
\]

For Property (iv) we assert that whenever $\isA(\var{np},\var{np}')$ holds and we have some instance
$p : \El_\NP^b(\var{np})$ satisfying an intersective adjective $\var{adj}:\IntAdj$, then $p$ also
satisfies $\var{adj}$ when regarded as an instance of $\var{np}'$:
\[
  \inferrule
  {\Gamma\vdash b, b': \Boundedness \\
   \Gamma\vdash \var{np} : \NP^b \\
   \Gamma\vdash \var{np}': \NP^{b'} \\
   \Gamma\vdash \var{isA} : \isA(\var{np},\var{np}') \\
   \Gamma\vdash \var{adj} : \IntAdj \\
   \Gamma\vdash p : \El_{\NP}^b(\var{np}) \\
   \Gamma\vdash \var{prf} : \Prff(\El_{\IA}(\var{adj})\ ((b,\var{np}),p))}
  {\Gamma\vdash \IARespectsIsA(\var{isA}, \var{adj}, p, \var{prf}) :
   \Prff(\El_{\IA}(\var{adj})\ ((b',\var{np}'), \El_\isA(\var{isA})\ p))}
\]

\begin{remark}
The above rule is the reason why we introduced $\isA$ in Section~\ref{sec:Subtyping in the nominal
domain} rather than simply modeling \emph{every $\var{np}$ is an $\var{np}'$} by the existence of a
function $\El_\NP^b(\var{np})\to\El_\NP^{b'}(\var{np}')$. It is much too general to say that
intersective adjectives respect arbitrary functions because there are too many definable functions
between the types of instances of noun phrases. Nor can we directly say that $\El_\NP^b(\var{np})$
is a ``subset'' of $\El_\NP^{b'}(\var{np}')$ in type theory.

For example, given $\john:\El_\NP^\U(\human)$ we can define a constant function $\lambda a.\john :
\El_\NP^\U(\cat)\to\El_\NP^\U(\human)$. If we defined $\isA(\var{np},\var{np}')$ as the type
$\El_\NP^\U(\cat)\to\El_\NP^\U(\human)$, then the rule for $\IARespectsIsA$ above would entail that
\emph{a black cat is a black human}.
\end{remark}

\begin{example}
As an illustration of Properties (iii) and (iv), we derive a function from instances of \emph{striped
(black cat)} to instances of \emph{striped cat}. To our framework we add:
\begin{mathpar}
  \inferrule
  { }
  {\Gamma\vdash \cat : \NP^\U}
  \and
  \inferrule
  { }
  {\Gamma\vdash \blk : \IntAdj}
  \and
  \inferrule
  { }
  {\Gamma\vdash \striped : \IntAdj}
\end{mathpar}

Following Example~\ref{ex:black cats formal}, we model \emph{black cat}, \emph{striped black cat},
and \emph{striped cat} as follows:
\begin{gather*}
  \blackCat \triangleq \sumNP_{p:\cat} \left(\El_{\IA}(\blk)\ ((\U,\cat),p)\right)
  : \NP^\U
  \\
  \textsf{stripedCat} \triangleq \sumNP_{p:\cat} \left(\El_{\IA}(\striped)\ ((\U,\cat),p)\right)
  : \NP^\U
  \\
  \textsf{stripedBlackCat} \triangleq \sumNP_{p:\blackCat}
  \left(\El_{\IA}(\striped)\ ((\U,\blackCat),p)\right)
  : \NP^\U
\end{gather*}

By the $\IANPIsNP$ rule, we have that every black cat is a cat:
\[
  \textsf{bcIsC} \triangleq \IANPIsNP(\cat,\blk) : \isA(\blackCat,\cat)
\]
Applying the $\IARespectsIsA$ rule to \emph{every black cat is a cat} and the adjective
\emph{striped}, we have that for any $p : \El_\NP^\U(\blackCat)$ and proof $\var{prf} :
\Prff(\El_{\IA}(\striped)\ ((\U,\blackCat),p))$ that $p$ is striped as a black cat, we also
have a proof that $\El_\isA(\textsf{bcIsC})\ p : \El_\NP^\U(\cat)$ is striped as a cat.
\[
  \IARespectsIsA(\textsf{bcIsC}, \striped, p, \var{prf}) :
   \Prff(\El_{\IA}(\striped)\ ((\U,\cat), \El_\isA(\textsf{bcIsC})\ p))
\]

To construct a function $\El_\NP^\U(\textsf{stripedBlackCat}) \to \El_\NP^\U(\textsf{stripedCat})$,
we must take as input a pair of some $p : \El_\NP^\U(\blackCat)$ and a proof $\var{prf} :
\Prff(\El_{\IA}(\striped)\ ((\U,\blackCat),p))$ that $p$ is striped as a black cat, and
produce as output a pair of some $p' : \El_\NP^\U(\cat)$ and a proof $\var{prf}\;' :
\Prff(\El_{\IA}(\striped)\ ((\U,\cat),p'))$ that $p'$ is striped as a cat. We take $p' \triangleq
\El_\isA(\textsf{bcIsC})\ p$, and we take $\var{prf}\;' \triangleq \IARespectsIsA(\textsf{bcIsC},
\striped, p, \var{prf})$ as described above.
\end{example}

Property (v)---that two \emph{black cats} are \emph{two black cats}---follows from several
properties of count noun phrases already stipulated in Section \ref{sec:Degrees and Units}, as shown
in the example below.

\begin{example}\label{ex:two black cats formal}
As an illustration of Property (v), we show how to derive one instance of \emph{two black cats} from
two instances of \emph{black cat}. Let us continue on from Example~\ref{ex:black cats formal}, in
which we defined \emph{black cat} as follows:
\[
  \blackCat \triangleq \sumNP_{p:\cat} \left(\El_{\IA}(\blk)\ ((\U,\cat),p)\right)
  : \NP^\U
\]
and showed that $(\tom,\tomIsBlack) : \El_\NP^\U(\blackCat)$ is a \emph{black cat}, where
$\tom:\El_{\NP}^\U(\cat)$ and $\tomIsBlack : \Prff(\El_{\IA}(\blk)\ ((\U,\cat),\tom))$. If we
further suppose
\begin{mathpar}
  \inferrule
  { }
  {\Gamma\vdash \felix : \El_{\NP}^\U(\cat)}
  \and
  \inferrule
  { }
  {\Gamma\vdash \felixIsBlack : \Prff(\El_{\IA}(\blk)\ ((\U,\cat),\felix))}
\end{mathpar}
then we have a second instance $(\felix,\felixIsBlack) : \El_\NP^\U(\blackCat)$ of a \emph{black
cat}.

If we model that \emph{cat} is a count noun phrase by adding the following rule:
\[
  \inferrule
  { }
  {\textsf{catIsCount} : \Prff(\isCount(\cat))}
\]
then by the closure of count noun phrases under $\Sigma$-types ($\SumIsCount$ from
Section~\ref{sec:Degrees and Units}), it follows that \emph{black cat} is also a count noun phrase:
\[
  \textsf{prf} \triangleq
  \SumIsCount(\cat,\textsf{catIsCount},\lambda p.\El_{\IA}(\blk)\ ((\U,\cat),p)) :
  \Prff(\isCount(\blackCat))
\]

Then, by the fact that a $\var{np}$ is a $\AmountOf(\var{np},\quantity, \natunits, 1)$ for any count
noun phrase $\var{np}$ ($\NPIsOneNP$ from Section~\ref{sec:Degrees and Units}), we have a function
$f$ that converts instances of \emph{black cat} into instances of \emph{one black cat}:
\begin{gather*}
  \textsf{isa} \triangleq
  \NPIsOneNP(\blackCat,\textsf{prf}) :
  \isA(\blackCat, \AmountOf(\blackCat,\quantity,\natunits,1))
  \\
  f \triangleq \El_\isA(\textsf{isa}) : \El_\NP^\U(\blackCat)\to
  \El_\NP^\B(\AmountOf(\blackCat,\quantity,\natunits,1))
\end{gather*}

Given two instances of \emph{one black cat} we easily obtain an instance of \emph{two black cats} by
$\oplus$:
\[
  f\ (\tom,\tomIsBlack) \oplus f\ (\felix,\felixIsBlack) : \AmountOf(\blackCat,\quantity,\natunits, 2)
\]
Note that our framework automatically determines that \emph{two black cats} is bounded: an
$\AmountOf$ an $\NP^\U$ (such as \emph{black cat}) is an $\NP^\B$.
\end{example}

Finally, for Property (vi)---two \emph{black cats} are a \emph{black} instance of \emph{two
cats}---we assert that $\oplus$ preserves intersective adjectives. Although the rule appears
intimidating, it simply states that if we have an instance $p$ of $m$ $\var{np}$s, a second instance
$q$ of $n$ of the same $\var{np}$ (in the same degree and units), and both $p,q$ satisfy some
intersective adjective $\var{adj}$, then $p\oplus q$ also satisfies $\var{adj}$:
\[
  \inferrule{
    \Gamma \vdash \var{np}:\NP^\U \\
    \Gamma \vdash d:\Degree \\
    \Gamma \vdash u:\Units(d) \\
    \Gamma \vdash m, n : \Nat \\\\
    \Gamma \vdash p:\El_{\NP}^\B(\AmountOf(\var{np},d,u,m)) \\
    \Gamma \vdash q:\El_{\NP}^\B(\AmountOf(\var{np},d,u,n)) \\\\
    \Gamma \vdash \var{adj} : \IntAdj \\
    \Gamma \vdash \var{prf} : \Prff(\El_\IA(\var{adj})\ ((\B, \AmountOf(\var{np},d,u,m)),p)) \\
    \Gamma \vdash \var{prf}\;' : \Prff(\El_\IA(\var{adj})\ ((\B, \AmountOf(\var{np},d,u,n)),q))
  }{
    \Gamma \vdash \oplusPreservesIA(\var{prf},\var{prf}\;')
    : \Prff(\El_\IA(\var{adj})\ ((\B, \AmountOf(\var{np},d,u,m+n)),p\oplus q))
  }
\]

\begin{example}\label{ex:black two cats formal}
To illustrate Property (vi), we continue on from Example~\ref{ex:two black cats formal}, sketching
how to obtain a \emph{black} instance of \emph{two cats} from the \emph{black cats} Tom and Felix.
As always, such an instance is a pair of an instance of \emph{two cats}
$\AmountOf(\cat,\quantity,\natunits, 2):\NP^\B$ along with a proof that the given instance is
\emph{black}.

Because we have stipulated that \emph{cat} is a count noun phrase, any \emph{cat} is \emph{one cat}:
\[
  \textsf{isa}' \triangleq \NPIsOneNP(\cat,\textsf{catIsCount}) :
  \isA(\cat,\AmountOf(\cat,\quantity,\natunits,1))
\]
In particular, $\tom : \El_\NP^\U(\cat)$ can be regarded as \emph{one cat}:
\[
  \El_\isA(\textsf{isa}')\ \tom : \El_\NP^\B(\AmountOf(\cat,\quantity,\natunits,1))
\]
and likewise with $\El_\isA(\textsf{isa}')\ \felix$. In addition, because intersective adjectives
respect $\isA$, we know that $\tom$ \emph{qua} \emph{one cat} is also \emph{black} (and likewise for
$\felix$):
\begin{gather*}
  \IARespectsIsA(\textsf{isa}', \blk, \tom, \tomIsBlack) :
  \\
  \Prff(\El_{\IA}(\blk)\ ((\B,\AmountOf(\cat,\quantity,\natunits,1)), \El_\isA(\textsf{isa}')\ \tom))
\end{gather*}

Putting the pieces together, we obtain an instance of \emph{two cats} as the sum of $\tom$
\emph{qua} \emph{one cat} and $\felix$ \emph{qua} \emph{one cat},
\[
  (\El_\isA(\textsf{isa}')\ \tom) \oplus (\El_\isA(\textsf{isa}')\ \felix) :
  \El_\NP^\B(\AmountOf(\cat,\quantity,\natunits, 2))
\]
and we obtain a proof that the above instance is \emph{black} by applying $\oplusPreservesIA$ to the
two previously constructed proofs that $\tom$ (resp., $\felix$) \emph{qua} \emph{one cat} are
\emph{black}.
\end{example}

\begin{remark}
	Property~(vi) should be read as a derived entailment about instances: from an instance of
	\emph{two black cats} we obtain an instance of \emph{two cats} together with a proof that it is
	\emph{black}. It is not meant to suggest that the English phrase \emph{two black cats} is composed by
	first forming \emph{two cats} and then applying \emph{black}; rather, \emph{two black cats} is formed
	by applying the quantity expression \emph{two} to the intersectively restricted noun phrase
	\emph{black cats}, as in Example~\ref{ex:two black cats formal}.
\end{remark}

%% file: sections/verbal.tex
Having now accounted for boundedness and unboundedness in the nominal domain, we turn our attention
to the main focus of our work: telicity, atelicity, and culminativity in the verbal domain. Our
treatment of telicity relies on a variation of event semantics in which we treat events themselves
as types, namely, the collections of their occurrences. This is different from previous
type-theoretic approaches to event semantics \citep{1994_Ranta_BOOK, 2015_Ranta_CONF, 2017_Luo_CONF,
2023_Cooper_BOOK}, where events are not treated as types but as either variables in the fashion of
neo-Davidsonian event semantics \citep{1990_Parsons_BOOK} or as proof objects of propositions.

Rather than having a single type universe of events, we follow \cite{2017_Luo_CONF} by considering a
dependent family of event types indexed by two semantic macroroles, which we call the actor and the
undergoer. In Section~\ref{sec:Preliminaries for event ontology}, we define types of actors and
undergoers, where the type of undergoers is further indexed by boundedness. In
Section~\ref{sec:Events(subsec)}, we define our types of events along with a family of types which
sends every event to its type of occurrences. In Section~\ref{sec:telic-atelic-events}, we show how
our dependent event types allow us to compositionally determine the (a)telicity of an event in terms
of the type of the event's undergoer.

In Section~\ref{sec:states}, we introduce the notion of states, the static counterpart to events. In
Section~\ref{sec:culminating}, we associate a resulting state to every telic event, and define
culminating events as telic events that, once they occur, obtain their resulting state. In
Section~\ref{sec:Adverbial modification} we discuss how our treatment of events as types allows us
to easily model adverbial modification along the lines of \cite{2015_Ranta_CONF}. Finally, in
Section~\ref{sec:Entailments} we discuss how to model various entailments between events not covered
in previous sections.

We note that our approach to event semantics, in addition to accounting for questions related to
telicity and culminativity, also provides an ontology of eventualities and their structure, an area
with little research in the context of type theory. (The only other work we are aware of is that of
\cite{2020_Corfield_BOOK}.) This treatment can be also seen as a type-theoretic analog of lexical
decomposition of verbs as studied in \cite{1979_Dowty_BOOK, 1998_Malka, 2005_Valin_BOOK}. The
benefit of our type-theoretic approach is that it provides us an avenue for developing a
set-theoretic semantics of our representation (see Section~\ref{sec:Future}), while
non-type-theoretic works on lexical decomposition do not provide any semantics at all.

Our event semantics preserves the main benefits of neo-Davidsonian event semantics (notably a
straightforward model of adverbial modification and its associated entailments) while avoiding its
main unappealing feature: the need for an existential quantifier in the representation of every
sentence. This is usually achieved by either including the existential quantifier in the lexical
semantics of verbs or using an existential closure rule, which is as ad hoc as the meaning
postulates for entailments involving adverbial modification that event semantics aims to overcome in
the first place. The necessity of existential quantifiers in neo-Davidsonian event semantics also
gives rise to the infamous event quantification problem \citep{2011_Winter_CONF, 2015_Groote_CONF,
2015_Champollion}, which does not arise in our case due to the absence of quantifiers. Although this
is not a unique feature of our approach, we also avoid the problem of non-compositionality that
arises in neo-Davidsonian event semantics, as discussed in \cite{1994_Ranta_BOOK, 2015_Ranta_CONF}.

\subsection{Preliminaries}\label{sec:Preliminaries for event ontology}

In our framework we will index the type of eventualities (and in particular events, which as
discussed in Remark~\ref{rem:eventuality} are dynamic eventualities) by two arguments, actors and
undergoers. As discussed in Section~\ref{sec:Relationship between telicity and undergoers}, an
\emph{actor} is the most agent-like participant in an event, whereas an \emph{undergoer} is the most
patient-like participant. In the sentence \emph{John ate apples}, for example, \emph{John} is the
actor and \emph{apples} are the undergoer.

One might be tempted to imagine that the types of actors and undergoers are both the type $\NP$, but
there are three issues with this. First, entities (Section~\ref{sec:ProperNouns}) such as
\emph{John} can be actors or undergoers but they are not noun phrases in our framework (see also
Remark~\ref{rem:NPterminology}). Second, some events (e.g., those described by intransitive verbs)
lack an actor or an undergoer, such as \emph{John died} (which lacks an actor) or \emph{John ran}
(which lacks an undergoer). Finally, we want to use type dependency to track the boundedness of
undergoers, because this determines the telicity of an event.

With these issues in mind, we declare the type of actors, $\Actor$, as a disjoint union of the type
of noun phrases $\NP^b$ (for any $b : \Boundedness$), the type of entities $\Entity$, and a ``dummy
actor'' $\actTm_\star$:
\begin{mathpar}
  \inferrule
  { }
  {\Gamma\vdash \Actor \type}
  \and
  \inferrule
  {\Gamma\vdash b : \Boundedness \\
   \Gamma\vdash \var{np} : \NP^b}
  {\Gamma\vdash \actTm_\NP(\var{np}) : \Actor}
  \and
  \inferrule
  {\Gamma\vdash e : \Entity}
  {\Gamma\vdash \actTm_\Entity(e) : \Actor}
  \and
  \inferrule
  { }
  {\Gamma\vdash \actTm_\star : \Actor}
\end{mathpar}

\begin{example}\label{ex:actors}
The following three sentences exemplify the three kinds of actors:

\ex.
\a. Three women popped the balloons. \label{3 women popped the balloons}
\b. John popped the balloons. \label{John popped the balloons}
\c. The balloon popped. \label{the balloons popped}

Assuming the relevant constants have been added to the theory,
\ref{3 women popped the balloons} has actor
\[
  \actTm_\NP(\AmountOf(\woman,\quantity,\natunits, 3)) : \Actor
\]
\ref{John popped the balloons} has actor $\actTm_\Entity((\U,\human),\john) : \Actor$, and
\ref{the balloons popped} has dummy actor $\actTm_\star : \Actor$.
\end{example}

As for undergoers, for any $b:\Boundedness$ we introduce the type $\Undergoer^b$ of ``undergoers
with boundedness $b$.'' Terms of type $\Undergoer^\B$ are \emph{bounded undergoers} and terms of
type $\Undergoer^\U$ are \emph{unbounded undergoers}.
\[
  \inferrule
  {\Gamma\vdash b:\Boundedness}
  {\Gamma\vdash \Undergoer^b \type}
\]

There are again three sources of undergoers---noun phrases, entities, and a dummy undergoer---but
now we must specify the boundedness of each kind of undergoer. $b$-noun phrases yield
$b$-undergoers, entities yield bounded undergoers, and the dummy is an unbounded undergoer:
\begin{mathpar}
  \\
  \inferrule
  {\Gamma\vdash b:\Boundedness \\
   \Gamma\vdash \var{np}:\NP^b}
  {\Gamma\vdash \undTm_\NP(\var{np}) : \Undergoer^b \type}
  \and
  \inferrule
  {\Gamma\vdash e : \Entity}
  {\Gamma\vdash \undTm_\Entity(e) : \Undergoer^\B}
  \and
  \inferrule
  { }
  {\Gamma\vdash \undTm_\star : \Undergoer^\U}
\end{mathpar}

\begin{example}\label{ex:undergoers}
Consider the following four sentences:

\ex.
\a. Three humans died. \label{3 humans died}
\b. John died. \label{John died}
\c. Balloons popped (for hours). \label{balloons popped for hours}
\d. John ran. \label{John ran}

Assuming the relevant constants have been added to the theory, \ref{3 humans died} has a bounded
undergoer $\undTm_\NP(\AmountOf(\human,\quantity,\natunits, 3)) : \Undergoer^\B$ because \emph{three
humans} is a bounded noun phrase; \ref{John died} has a bounded undergoer
$\undTm_\Entity((\U,\human),\john) : \Undergoer^\B$ because all entities yield bounded undergoers.
Unbounded undergoers include unbounded noun phrases, such as $\undTm_\NP(\balloon) : \Undergoer^\U$
in \ref{balloons popped for hours}, and the dummy undergoer $\undTm_\star:\Undergoer^\U$ in
\ref{John ran}.
\end{example}

As with $\NP$ in Section~\ref{sec:UniversesOfNP}, we can define a type of undergoers with arbitrary
boundedness:
\[
  \inferrule
  { }
  {\Gamma\vdash \Undergoer \triangleq \textstyle\sum_{b:\Boundedness} \Undergoer^b \type}
\]

Because actors and undergoers exist purely to index the type of events, some nuances of their
definitions are best illustrated after we explain how our framework models events. We will explain
the purpose of dummy actors and undergoers in Section~\ref{sec:Events(subsec)}; in
Section~\ref{sec:telic-atelic-events}, we will explain why entity undergoers are bounded and the
dummy undergoer is unbounded.

\subsection{Events and entailments}\label{sec:Events(subsec)}

With the types of actors and undergoers in hand, we now declare the type of events
$\Evt^b(a,\var{und})$ with actor $a:\Actor$ and $b$-undergoer $\var{und}:\Undergoer^b$ (where
$b:\Boundedness$).
\[
\inferrule
    {\Gamma\vdash b:\Boundedness \\
     \Gamma\vdash a:\Actor \\
     \Gamma\vdash \var{und}:\Undergoer^b}
    {\Gamma\vdash \Evt^b(a,\var{und}) \type}
\]
Examples of events include the sentences in Examples~\ref{ex:actors}~and~\ref{ex:undergoers}.

In the same way that we associate to every noun phrase $\var{np}:\NP^b$ a type $\El_\NP^b(\var{np})$
of its instances, we will associate to every event $\var{evt}: \Evt^b(a,\var{und})$ a type
$\El_\Evt(e)$ of its \emph{occurrences}.
\[
  \inferrule
  {\Gamma\vdash b:\Boundedness \\
   \Gamma\vdash a:\Actor \\
   \Gamma\vdash \var{und}:\Undergoer^b \\
   \Gamma\vdash \var{evt}:\Evt^b(a,\var{und})}
  {\Gamma\vdash \El_\Evt(\var{evt}) \type}
\]

The type $\El_\Evt(\var{evt})$ serves two important roles in our framework. First, it gives us a way
to represent whether an event $\var{evt}:\Evt^b(a,\var{und})$ has occurred, namely that
$\El_\Evt(\var{evt})$ is inhabited (has at least one element). Second, treating events as types
rather than merely propositions will allow us in Section~\ref{sec:Adverbial modification} to account
for adverbial modification in the spirit of what is proposed in \cite{2015_Ranta_CONF}.

\begin{example}\label{ex:eat(john,apple) occurrences of events}
To model the event \emph{John ate apples}, our framework must include at least the following
constants previously considered in Section~\ref{sec:Nominal}:
\begin{mathpar}
  \inferrule
  { }
  {\Gamma \vdash \human:\NP^\U}
  \and
  \inferrule
  { }
  {\Gamma\vdash \john : \El_\NP^\U(\human)}
  \and
  \inferrule
  { }
  {\Gamma\vdash \apple : \NP^\U}
\end{mathpar}

We can model the transitive verb \emph{eat} as a dependent product of events, as follows:
\[
  \inferrule
  { }
  {\Gamma\vdash \eat: (a:\Actor)\to (u:\Undergoer)\to \Evt^{\fst(u)}(a,\snd(u))}
\]

By the rules for actors and undergoers in Section~\ref{sec:Preliminaries for event ontology}, we
have
\begin{gather*}
  \john_\Actor \triangleq \actTm_\Entity((\U,\human),\john) : \Actor \\
  \undTm_\NP(\apple) : \Undergoer^\U
\end{gather*}
and thus the term
\[
  \textsf{jaa} \triangleq \eat\ \john_\Actor\ (\U , \undTm_\NP(\apple))
  : \Evt^\U(\john_\Actor, \undTm_\NP(\apple))
\]
represents the event \emph{John ate apples}. The type $\El_\Evt(\textsf{jaa})$ then represents the
collection of instantiations or occurrences of the event of John eating apples, i.e., various times
and places when a particular John ate apples.
\end{example}

The fact that every event is a term of type $\Evt^b(a,\var{und})$ seems to suggest that every event
has both an actor and an undergoer, but of course sentences in natural language often lack one of
these, either because the event inherently has no actor or undergoer, or because the actor or undergoer is left implicit. In the former case, we
model the absence of an actor or undergoer by the dummies $\actTm_\star : \Actor$ and $\undTm_\star
: \Undergoer^\U$.

\begin{example}
Consider the sentences \emph{John died} (Example~\ref{ex:actors}), which has no actor, and
\emph{John ran} (Example~\ref{ex:undergoers}), which has no undergoer. Because the verb \emph{die}
never has an actor, we model it in our framework by the following dependent product of events, which
can have any undergoer but always the dummy actor $\actTm_\star$:
\[
  \inferrule
  { }
  {\Gamma\vdash \die : (u:\Undergoer)\to \Evt^{\fst(u)}(\actTm_\star,\snd(u))}
\]
Similarly, the verb \emph{run} can have any actor but always the dummy undergoer $\undTm_\star$:
\[
  \inferrule
  { }
  {\Gamma\vdash \run : (a:\Actor)\to \Evt^\U(a,\undTm_\star)}
\]

We thus model the sentences \emph{John died} and \emph{John ran} as the following events:
\begin{gather*}
  \die\ (\B,\undTm_\Entity((\U,\human),\john))
  : \Evt^\B(\actTm_\star, \undTm_\Entity((\U,\human),\john))
  \\
  \run\ (\actTm_\Entity((\U,\human),\john))
  : \Evt^\U(\actTm_\Entity((\U,\human),\john), \undTm_\star)
\end{gather*}
\end{example}

\begin{remark}\label{rem:no dummies}
It is possible to avoid dummy actors and undergoers by adding primitive types of events that have no
actor (resp., undergoer), and then defining the type of all events as the disjoint union of those
with specified actor and undergoer, those with specified actor and no undergoer, and those with no
actor and specified undergoer. Such a treatment would be substantially more complicated for
apparently little gain. Moreover, as we will remark in Example~\ref{ex:pop balloon}, dummy actors
and undergoers give us a convenient way to model verbs such as \emph{pop} that can be used both transitively
and intransitively.
\end{remark}

The other reason why a sentence might lack an actor or undergoer is that it might be left implicit
by the sentence despite existing. Such is the case with the undergoer in the sentence \emph{John
ate}: it must be that John ate \emph{something}, but that thing is left unspecified.

Actors may likewise be left unspecified. Verbs that participate in the causative--anticausative
alternation illustrate this point. Depending on context, the sentence \emph{The balloon popped}
supports two alternative event construals: an internally-caused event with no actor, or an
externally-caused event whose actor is implicit. Because there is no overt actor, the syntax favors
the no-actor construal; however, world knowledge often pragmatically enriches the interpretation
toward an implicit external cause such as environmental conditions or a sentient individual. We do not attribute these two construals to a lexical semantic ambiguity for \emph{pop}, but
rather to contextual/pragmatic inference.

This perspective also explains why a speaker who has just watched John puncture the balloon may
still felicitously say \emph{The balloon popped}: the utterance leaves the actor unexpressed, while
the context supplies a specific value for the implicit actor, similar to a specific-indefinite
reading of \emph{Someone popped the balloon} where the \emph{someone} is known to be John.

We may account for unspecified actors or undergoers using $\Sigma$-types. For example, we define the
type of events $\Evt_\A(a)$ with a specified actor $a:\Actor$ but unspecified undergoer as
consisting of pairs of an undergoer $u:\Undergoer$ and an event with actor $a$ and undergoer $u$. We
likewise define the type of events $\Evt_\Patient(\var{und})$ with a specified undergoer
$\var{und}:\Undergoer^b$ but unspecified actor, and even the type of events $\Evt$ whose actor and
undergoer are both unspecified.
\begin{mathpar}
  \inferrule
  {\Gamma\vdash a:\Actor}
  {\Gamma\vdash \Evt_\A(a) \triangleq \textstyle\sum_{u:\Undergoer} \Evt^{\fst(u)}(a,\snd(u)) \type}
  \and
  \inferrule
  {\Gamma\vdash b:\Boundedness \\
   \Gamma\vdash \var{und}:\Undergoer^b}
  {\Gamma\vdash \Evt_\Patient^b(\var{und}) \triangleq
   \textstyle\sum_{a:\Actor} \Evt^{b}(a,\var{und}) \type}
  \and
  \inferrule
  { }
  {\Gamma\vdash \Evt \triangleq
   \textstyle\sum_{a:\Actor}\textstyle\sum_{u:\Undergoer} \Evt^{\fst(u)}(a,\snd(u)) \type}
\end{mathpar}
Note that an unspecified actor/undergoer \emph{may} be, but is not required to be, the dummy
actor/undergoer, so these types are \emph{not} those discussed in Remark~\ref{rem:no dummies}. (See
also Example~\ref{ex:pop balloon} and Remark~\ref{rem:dummy disambiguation} for more discussion of
the distinction between unspecified and dummy actors/undergoers.)

The rules for $\Sigma$-types let us define functions that suppress an actor and/or undergoer, e.g.
\[
  \lambda\var{evt}. ((b,\var{und}),\var{evt}) : \Evt^b(a,\var{und}) \to \Evt_\A(a)
\]
Furthermore, by postcomposing the second projection
\[
  \snd : (\var{evt}:\Evt_\A(a)) \to \Evt^{\fst(\fst(\var{evt}))}(a,\snd(\fst(\var{evt})))
\]
by the previously-asserted $\El_\Evt$, we can associate to each $\var{evt}_\A:\Evt_\A(a)$ the type
$\El_{\Evt_\A}(\var{evt}_\A)$
\[
  \inferrule
  {\Gamma\vdash a:\Actor \\
   \Gamma\vdash \var{evt}_\A : \Evt_\A(a)}
  {\Gamma\vdash \El_{\Evt_\A}(\var{evt}_\A) \triangleq \El_\Evt(\snd(\var{evt}_\A)) \type}
\]
of its occurrences. We likewise define $\El_{\Evt_\Patient^b}(\var{evt}_\Patient) \triangleq
\El_\Evt(\snd(\var{evt}_\Patient))$.

\begin{example}\label{ex:john ate}
Continuing on from Example~\ref{ex:eat(john,apple) occurrences of events}, we can model the event
\emph{John ate} as a term of type $\Evt_{\A}(\john_\Actor)$. Such terms are a pair $(u,\var{evt})$
of an undergoer $u:\Undergoer$ and an event $\var{evt} : \Evt^{\fst(u)}(\john_\Actor,\snd(u))$
involving the entity John and the undergoer $u$. (Here we are ignoring the fact that there are some
restrictions on what type of undergoer the verb \emph{eat} can take, since modeling this kind of
selectional restriction is not the focus of this paper.)

Because the type of occurrences $\El_{\Evt_\A}(u,\var{evt})$ of an event-with-unspecified-undergoer
$(u,\var{evt}) : \Evt_\A(\john_\Actor)$ is defined as the type of occurrences $\El_\Evt(\var{evt})$
of its underlying event-with-specified-undergoer, the collection of occurrences of \emph{John ate}
is precisely the collection of occurrences in which \emph{John ate something}, that is, the
collection of occurrences that this particular John ate some undergoer.
\end{example}

Rephrasing the observation at the end of Example~\ref{ex:john ate}, if it is true that \emph{John
ate apples}, then it is also true that \emph{John ate}, but the converse does not necessarily hold.
Thus we have automatically a one-way entailment between the sentences involving verbs like
\emph{eat}, which may describe events-with-unspecified-undergoers.

\begin{example}\label{ex:pop balloon}
We now consider an example of the causal-noncausal alternation, namely \emph{John popped balloons}
vs.\ \emph{Balloons popped}, and the associated entailment relations. As discussed earlier in this
section, a sentence like \emph{Balloons popped} may be interpreted as describing an event either
with no actor or with an unspecified actor. In the former case, the sentence would be interpreted as
\emph{Balloons popped by themselves, without anything or anyone causing that}, while in the latter
case the sentence would be interpreted as \emph{Balloons popped, and in fact, someone/something
popped them} (where \emph{someone/something} is used in the specific sense---i.e., it is known to
the utterer). Accordingly, the entailment \emph{John popped balloons} $\Rightarrow$ \emph{Balloons
popped} should not hold in the former case (since John popping balloons contradicts balloons popping
by themselves), and it should only hold in the latter case if the ``someone'' that the utterer has
in mind is John.

We now show how our theory accounts for this. To model these events we postulate:
\begin{mathpar}
  \inferrule
  { }
  {\Gamma \vdash \human:\NP^\U}
  \and
  \inferrule
  { }
  {\Gamma\vdash \john : \El_\NP^\U(\human)}
  \and
  \inferrule
  { }
  {\Gamma \vdash \balloon:\NP^\U}
  \and
  \inferrule
  { }
  {\Gamma\vdash \pop: (a:\Actor)\to (u:\Undergoer)\to \Evt^{\fst(u)}(a,\snd(u))}
\end{mathpar}

Because of our treatment of dummy undergoers, $\pop$ represents both the transitive and intransitive
versions of the English verb \emph{pop}. Again writing
\[
  \john_\Actor \triangleq \actTm_\Entity((\U,\human),\john) : \Actor
\]
we can represent \emph{John popped balloons} as
\[
  \textsf{evt}_1 \triangleq
  \pop\ \john_\Actor \ (\U,\undTm_\NP(\balloon))
  : \Evt^\U(\john_\Actor,\undTm_\NP(\balloon))
\]
we can represent \emph{Balloons popped (in fact, John popped them)}  as
\[
  \textsf{evt}_2 \triangleq
  (\john_\Actor , \pop\ \john_\Actor \ (\U,\undTm_\NP(\balloon)))
  : \Evt^\U_\Patient(\undTm_\NP(\balloon))
\]
and finally, we can represent \emph{Balloons popped by themselves, without anything or anyone
causing that} as
\[
  \textsf{evt}_3 \triangleq
  (\actTm_\star , \pop\ \actTm_\star\ (\U,\undTm_\NP(\balloon)))
  : \Evt^\U_\Patient(\undTm_\NP(\balloon))
\]

Note that $\textsf{evt}_2$ and $\textsf{evt}_3$ have the same type
$\Evt^\U_\Patient(\undTm_\NP(\balloon))$ because they are both events with unspecified actor and
balloon undergoer, but the actor in the first case is \emph{John} and the actor in the second case
is the dummy actor.

In our framework, we can express the entailment \emph{John popped balloons} $\Rightarrow$
\emph{Balloons popped (in fact, John popped them)} by a function of type
$\El_\Evt(\textsf{evt}_1)\to\El_{\Evt^\U_\Patient}(\textsf{evt}_2)$. (Such functions witness an
entailment by the standard propositions as types reading; indeed, if there is at least one
occurrence of $\textsf{evt}_1$, then via such a function there is at least one occurrence of
$\textsf{evt}_2$.)

In this case, because \emph{John} is the suppressed actor in $\textsf{evt}_2$, the two types in the
implication are identical once we expand definitions, so the identity function suffices to establish
the implication, as well as the reverse implication $\El_{\Evt^\U_\Patient}(\textsf{evt}_2)\to
\El_\Evt(\textsf{evt}_1)$.

However, the entailment \emph{John popped balloons} $\Rightarrow$ \emph{Balloons popped by
themselves, without anything or anyone causing that},
$\El_\Evt(\textsf{evt}_1)\to\El_{\Evt^\U_\Patient}(\textsf{evt}_3)$, does not hold in our framework,
because there is no relationship between these two types. Nor do we have \emph{John popped balloons}
$\Rightarrow$ \emph{Balloons popped (in fact, Mary popped them)}.
\end{example}

\begin{remark}\label{rem:dummy disambiguation}
How do we determine whether the sentence \emph{balloons popped} means \emph{someone popped balloons}
or \emph{balloons popped by themselves}---that is, whether the sentence has a dummy undergoer or a
suppressed non-dummy undergoer? As in Remark~\ref{rem:several}, our answer is that the ambiguity of
who or what popped the balloons must be resolved, since in general any sentence must be
disambiguated prior to modeling.
\end{remark}


\subsection{Telic and atelic events}\label{sec:telic-atelic-events}

At this point it is simple to define the types of telic and atelic events. Recall from
Section~\ref{sec:bg-telicity} that telic events are events with inherent endpoints (or equivalently,
a naturally associated resulting state) such as \emph{John ate the soup}.  As discussed in
Section~\ref{sec:Relationship between telicity and undergoers}, we are restricting our attention to
cases in which inherent endpoints come from undergoers (not adjuncts), so we say that an event is
telic (resp., atelic) if and only if its undergoer is bounded (resp., unbounded).

Because our type of events already tracks the boundedness of its undergoer, we can simply define the
type of telic events $\Tel(a,\var{und})$ with actor $a:\Actor$ and bounded undergoer
$\var{und}:\Undergoer^\B$ as the type of events $\Evt^\B(a,\var{und})$, and analogously for the type
of atelic events:
\begin{mathpar}
\inferrule
  {\Gamma\vdash a:\Actor \\
   \Gamma\vdash \var{und}:\Undergoer^\B}
  {\Gamma\vdash \Tel(a,\var{und}) \triangleq \Evt^\B(a,\var{und}) \type}
\and
\inferrule
  {\Gamma\vdash a:\Actor \\
   \Gamma\vdash \var{und}:\Undergoer^\U}
  {\Gamma\vdash \Atel(a,\var{und}) \triangleq \Evt^\U(a,\var{und}) \type}
\end{mathpar}
Using $\Sigma$-types, we can once again define types of telic/atelic events where one or none of
actor or undergoer are specified: $\Tel_\A(a)$, $\Tel_\Patient(\var{und})$, $\Tel$, $\Atel_\A(a)$,
$\Atel_\Patient(\var{und})$, and $\Atel$.

Recall from Section~\ref{sec:Preliminaries for event ontology} that there are three kinds of
undergoers: noun phrases, whose boundedness is that of the underlying noun phrase; entities, which
are always bounded; and the dummy undergoer, which is always unbounded. We now reconsider each of
these undergoers in turn.

\begin{example}\label{ex:john ate apples telicity}
Revisiting \emph{John ate apples} from Example~\ref{ex:eat(john,apple) occurrences of events},
recall that \emph{eat} has type:
\[
  \inferrule
  { }
  {\Gamma\vdash \eat: (a:\Actor)\to (u:\Undergoer)\to \Evt^{\fst(u)}(a,\snd(u))}
\]
and that \emph{apple} is an unbounded noun phrase and thus $\undTm_\NP(\apple) : \Undergoer^\U$. The
event \emph{John ate apples} is thus modeled as an atelic event
\[
  \eat\ \john_\Actor\ (\U , \undTm_\NP(\apple))
  : \Atel(\john_\Actor, \undTm_\NP(\apple))
\]
as predicted by the \emph{in}-adverbial test described in Section~\ref{sec:Relationship between
telicity and undergoers}, \ref{ex:TelicityCorrelatesWithBoundednessApples}.

On the other hand, \emph{three apples} is a bounded noun phrase
\[
  \textsf{threeApples} \triangleq \AmountOf(\apple,\quantity,\natunits, 3):\NP^\B
\]
by Section~\ref{sec:Degrees and Units}, so the event \emph{John ate three apples} is modeled as a
telic event
\[
  \eat\ \john_\Actor\ (\B , \undTm_\NP(\textsf{threeApples}))
  : \Tel(\john_\Actor, \undTm_\NP(\textsf{threeApples}))
\]
again as predicted by the \emph{in}-adverbial test described in Section~\ref{sec:Relationship
between telicity and undergoers}, \ref{ex:TelicityCorrelatesWithBoundednessThreeApples}.

What about entity undergoers? Recalling that an entity is an instance of a noun phrase, the events
\emph{John ate a particular apple} and \emph{John ate a particular three apples} both have an entity
as undergoer. Supposing that $a_1,a_2,a_3 : \El_{\NP}^\U(\apple)$ are three particular apples, then
\[
  \undTm_\Entity((\U, \apple), a_1) : \Undergoer^\B
\]
is the undergoer representing the \emph{particular apple} $a_1$, and
\[
  \undTm_\Entity((\B, \textsf{threeApples}), f(a_1) \oplus f(a_2) \oplus f(a_3)) : \Undergoer^\B
\]
is the undergoer representing the \emph{particular three apples} $a_1,a_2,a_3$, where $f$ in the
above term is a function which converts an apple into a \emph{one apple}, along the lines of
Example~\ref{ex:john+mary}.

Crucially, although \emph{apple} is unbounded and \emph{three apples} is bounded, both \emph{a
particular apple} and \emph{a particular three apples} are bounded, because all entities are bounded
undergoers. Thus our framework correctly detects both \emph{John ate a particular apple} and
\emph{John ate a particular three apples} as telic events.
\end{example}

\begin{remark}
	Note that our treatment of bare plurals differs from the mereological framework
	\citep{2016_Champollion_Chapter} widely used to model plurality. In that framework, the bare plural
	\emph{apples} denotes the set of all mereological sums of apples \cite[44]{2017_Champollion_BOOK},
	and \emph{John ate apples} is true just in case there exists some sum of apples that John ate.
	Such treatments are inadequate in our setting: representing this existential commitment amounts to selecting a
	particular sum (an entity) as an undergoer, and since entity undergoers are classified
	as bounded in our event ontology, this incorrectly forces \emph{John ate apples} to be telic. 
\end{remark}

Finally, the dummy undergoer $\undTm_\star : \Undergoer^\U$ is unbounded because, at least in
English, sentences with unspecified undergoers seem to always be atelic. For example, the
unspecified undergoer in \emph{John sang} seems to refer to \emph{songs} rather than \emph{a song};
\emph{John sang} is atelic (by the \emph{in}-adverbial diagnostic), and if the unspecified undergoer
was \emph{a song}, then it should have been telic. Moreover, verbs that never take undergoers, such
as \emph{John ran} \ref{John ran}, seem to always be atelic---at least if one does not add adjuncts,
which are beyond the scope of our analysis.

\subsection{States}\label{sec:states}

So far our discussion on the verbal domain has focused on events, such as \emph{The balloon popped},
which express dynamic eventualities. We now turn our attention to static eventualities, such as
\emph{The balloon is popped}, which we will call \emph{states}. We introduce states for two reasons.
First, one of our objectives is to build a reasonable ontology of eventualities, and any such
ontology should include states. Second, in Section~\ref{sec:culminating}, we will characterize
culminating events as telic events that attain their resulting state.

Being the static analog of events, the type of states mirrors the type of events defined in
Section~\ref{sec:Events(subsec)}: for any actor $a:\Actor$ and $b$-undergoer
$\var{und}:\Undergoer^b$ (where $b:\Boundedness$), we have a type $\State^b(a,\var{und})$ of states
involving that actor and undergoer.
\[
\inferrule
    {\Gamma\vdash b:\Boundedness \\
     \Gamma\vdash a:\Actor \\
     \Gamma\vdash \var{und}:\Undergoer^b}
    {\Gamma\vdash \State^b(a,\var{und}) \type}
\]

For any state $s:\State^b(a,\var{und})$, we can consider the proposition $\El_\State(s) : \Prop$
that the state $s$ holds. Note that unlike events $\var{evt} : \Evt^b(a,\var{und})$, which give rise
to a type of occurrences $\El_\Evt(\var{evt})$, a state only gives rise to a proposition; this is
because we do not consider a state to be able to hold in more than one way.
\[
  \inferrule
  {\Gamma\vdash b:\Boundedness \\
   \Gamma\vdash a:\Actor \\
   \Gamma\vdash \var{und}:\Undergoer^b \\
   \Gamma\vdash s:\State^b(a,\var{und})}
  {\Gamma\vdash \El_\State(s) : \Prop}
\]

Using $\Sigma$-types, we can also define types of states in which one or none of actor or undergoer
are specified---$\State_\A(a)$, $\State_\Patient^b(\var{und})$, and $\State$---each of which has an
underlying state of type $\State^b(a,\var{und})$ and is therefore also associated to a proposition.
As with events, suppressed arguments may or may not be dummies.

For example, the sentence \emph{John loves Mary} expresses a state whose undergoer is \emph{Mary}
and whose actor is \emph{John}; the sentence \emph{Mary is loved} expresses a state whose undergoer
is \emph{Mary} and whose actor is suppressed (but possibly \emph{John}). As with events (\emph{John
ran}), some states may be required to have a dummy argument: the sentence \emph{The balloon is
popped} is a state whose undergoer is \emph{the balloon} and whose actor must be the dummy.

\subsection{Culminating events}\label{sec:culminating}

We finally turn our attention to culminativity. Recalling that telic events are events that have a
naturally associated resulting state, we define culminating events as telic events that obtain their
resulting state. Note that it is not sensible to even ask whether an atelic event is culminating,
because there is no resulting state to be obtained. Our definition is in line with previous work
that mentions culminativity, such as \cite{Moens1988}. In other works on lexical semantics that
equate culminating events with telic events, such as \cite{1979_Dowty_BOOK}, the defining property
of telic (for us, culminating) events is that they entail the resulting state.

We note that atelic events \emph{can} involve a change in state; they simply do not have a
\emph{resulting state}. For instance, the event \emph{John popped balloons} involves a change of
state, since some balloons are being popped, but it has no resulting state and is thus atelic. To
see that it has no resulting state, observe that it would be strange to continue \emph{\dots and he
finished popping them all}; this is because \emph{finish} refers to a telos, and \emph{John popped
balloons} doesn't involve one. In contrast, the telic event \emph{John popped three balloons} has a
clear resulting state, namely that the three balloons are popped, and in fact the event culminates
because it reaches that state.

We start by associating to every telic event a resulting state.
\[
  \inferrule
  {\Gamma\vdash a:\Actor \\
   \Gamma\vdash \var{und}:\Undergoer^\B \\
   \Gamma\vdash \var{evt}:\Tel(a,\var{und})}
  {\Gamma\vdash \Result(\var{evt}):\State^\B(\actTm_\star,\var{und})}
\]

Note that the resulting state of an event with actor $a$ and undergoer $\var{und}$ involves the same
undergoer but always the dummy actor $\actTm_\star$. Indeed, sentences describing the resulting
state of some event seem to always have the property that they do not have a counterpart in which an
actor is specified. On the other hand, not every $s:\State^\B(\actTm_\star,\var{und})$ is the
resulting state of a telic event; counterexamples include \emph{The apple is tasty} and \emph{John
is tired}.

With the notion of resulting state in hand, we can now formally define the culminativity of a telic
event as the proposition that if the event occurs, then its resulting state holds.
\begin{mathpar}
  \inferrule
  {\Gamma\vdash a : \Actor \\
   \Gamma\vdash \var{und} : \Undergoer^\B \\
   \Gamma\vdash \var{evt} : \Tel(a,\var{und})}
  {\Gamma\vdash \isCul(\var{evt}) : \Prop}
  \and
  \inferrule
  {\Gamma\vdash a : \Actor \\
   \Gamma\vdash \var{und} : \Undergoer^\B \\
   \Gamma\vdash \var{evt} : \Tel(a,\var{und})}
  {\Gamma\vdash \Prff(\isCul(\var{evt}))\triangleq
   \El_\Evt(\var{evt}) \to \Prff(\El_\State(\Result(\var{evt}))) \type}
\end{mathpar}
Depending on which formal definition of $\Prop$ we use, this definition of $\isCul$ may be
self-evidently a proposition (because it is of the form $X\to\Prff(Y)$ where $Y$ is a proposition)
or we may need to prove it. In our Agda formalization, we define $\Prop$ as the type of homotopy
propositions \citep{HoTT13} and use function extensionality to prove that $\isCul$ is an homotopy
proposition.

Finally, for any actor $a : \Actor$ and bounded undergoer $\var{und} : \Undergoer^\B$, we define
culminating events $\Cul(a,\var{und})$ as pairs of a telic event with a proof that it culminates:
\[
  \inferrule
  {\Gamma \vdash a : \Actor \\
   \Gamma \vdash \var{und} : \Undergoer^\B}
  {\Gamma \vdash \Cul(a,\var{und}) \triangleq
   \textstyle\sum_{\var{evt}:\Tel(a,\var{und})} \Prff(\isCul(\var{evt})) \type}
\]
We can again define the types of culminating events in which one or none of actor or undergoer are
specified---$\Cul_\A(a)$, $\Cul_\Patient(\var{und})$, and $\Cul$---and associate to each of these
types the type of its occurrences (namely, the type of occurrences of the underlying telic event).
Note that the undergoer of a culminating event must be bounded, because telic events are events with
a bounded undergoer. In particular, the undergoer cannot be the dummy undergoer.

\begin{example}\label{ex:pop culminating}
Consider the sentence \emph{John popped three balloons}, a telic and culminating event whose
resulting state is \emph{three balloons are popped}. The general setup is the same as in
Example~\ref{ex:pop balloon}. We postulate:
\begin{mathpar}
  \inferrule
  { }
  {\Gamma \vdash \human:\NP^\U}
  \and
  \inferrule
  { }
  {\Gamma\vdash \john : \El_\NP^\U(\human)}
  \and
  \inferrule
  { }
  {\Gamma \vdash \balloon:\NP^\U}
  \and
  \inferrule
  { }
  {\Gamma\vdash \pop: (a:\Actor)\to (u:\Undergoer)\to \Evt^{\fst(u)}(a,\snd(u))}
\end{mathpar}

From these we can define the actor \emph{John}, the undergoer \emph{three balloons} (which is
automatically determined by the framework to be bounded), and the (automatically determined to be)
telic event $\textsf{evt}_1$ representing \emph{John popped three balloons}.
\begin{gather*}
  \john_\Actor \triangleq \actTm_\Entity((\U,\human),\john) : \Actor
  \\
  \textsf{threeBalloons}_\Undergoer \triangleq
  \undTm_\NP(\AmountOf(\balloon,\quantity,\natunits,3)):\Undergoer^\B
  \\
  \textsf{evt}_1 \triangleq
  \pop\ \john_\Actor\ (\B , \textsf{threeBalloons}_\Undergoer)
  : \Tel(\john_\Actor,\textsf{threeBalloons}_\Undergoer)
\end{gather*}

To model the culminativity of \emph{pop}, we add three further postulates: (1) an undergoer-indexed
family of states, $\popped$, representing the static predicate \emph{be popped}; (2) an equation
stating that the $\Result$ of the event of $\pop$ping a bounded undergoer is the state of being
$\popped$; and (3) the axiom $\textsf{popC}$ that any occurrence of the event of $\pop$ping a
bounded undergoer causes the state of being $\popped$ to hold.
\begin{mathpar}
  \inferrule
  { }
  {\Gamma\vdash \popped : (\var{und}:\Undergoer^\B)\to\State^\B(\actTm_\star,\var{und})}
  \and
  \inferrule
  {\Gamma\vdash a : \Actor \\
   \Gamma\vdash \var{und} : \Undergoer^\B}
  {\Gamma\vdash \Result(\pop\ a\ (\B,\var{und})) \triangleq
   \popped\ \var{und} : \State^\B(\actTm_\star,\var{und})}
  \and
  \inferrule
  { }
  {\Gamma\vdash \textsf{popC} :
   (a:\Actor)\to (\var{und}:\Undergoer^\B)\to
   \El_\Evt(\pop\ a\ (\B,\var{und}))\to \Prff(\El_\State(\popped\ \var{und}))}
\end{mathpar}
Note that all three of these postulates refer only to bounded undergoers.

From these postulates we can show that $\pop$ is culminating when its undergoer is bounded. We can
express this by giving a more refined type to $\pop$ restricted to bounded undergoers; the following
term type-checks straightforwardly by expanding definitions.
\[
  \pop^\B \triangleq \lambda a. \lambda\var{und}. (\pop\ a\ \var{und} , \textsf{popC}\ a\ \var{und})
  : (a : \Actor)\to (\var{und} : \Undergoer^\B)\to \Cul(a,\var{und})
\]
Using $\pop^\B$, we can easily show that the events \emph{John popped three balloons} and
\emph{three balloons popped by themselves} are both culminating:
\begin{gather*}
  \textsf{evt}_1' \triangleq
  \pop^\B\ \john_\Actor\ \textsf{threeBalloons}_\Undergoer
  : \Cul(\john_\Actor,\textsf{threeBalloons}_\Undergoer)
  \\
  \textsf{evt}_2 \triangleq
  \pop^\B\ \actTm_\star\ \textsf{threeBalloons}_\Undergoer
  : \Cul(\actTm_\star,\textsf{threeBalloons}_\Undergoer)
\end{gather*}

Moreover, we can show that if the event \emph{John popped three balloons} occurs, this entails the
state that \emph{three balloons are popped}:
\[
  \snd(\textsf{evt}_1') :
  \El_\Evt(\textsf{evt}_1)\to\Prff(\El_\State(\popped\ \textsf{threeBalloons}_\Undergoer))
\]
\end{example}

\begin{example}\label{ex:type of pop}
English verbs like \emph{pop} and \emph{eat} are the most common kind of verbs in English: they
yield telic and culminating events when their undergoer is bounded and atelic events otherwise. In
Example~\ref{ex:pop culminating} we saw that we can capture the culminativity of bounded \emph{pop}
in the type of $\pop^\B$. In this example, we show that dependent type theory can capture both the
culminativity of bounded \emph{pop} and the atelicity of unbounded \emph{pop} in a single type. This
will require some more advanced features of dependent type theory, namely dependent pattern matching
\citep{Coquand92} and type universes $\Univ$.

Using dependent pattern matching, we can define a family of types $\CulOrAtel$ by cases on the
boundedness $b$ of an undergoer $u=(b,\var{und}):\Undergoer$. Specifically, we define $\CulOrAtel\
a\ (b,\var{und})$ to be $\Cul(a,\var{und})$ when $b=\B$ and $\Atel(a,\var{und})$ when $b=\U$:
\begin{gather*}
\CulOrAtel\ : (a : \Actor) \to (u : \Undergoer) \to \Univ \\
\CulOrAtel\ a\ (\B , \var{und}) \triangleq \Cul(a,\var{und}) \\
\CulOrAtel\ a\ (\U , \var{und}) \triangleq \Atel(a,\var{und})
\end{gather*}

Then, we define by dependent pattern matching a function $\pop'$ which for any actor $a:\Actor$ and
undergoer $u:\Undergoer$ produces a term of type $\CulOrAtel\ a\ u$:
\begin{gather*}
\pop'\ : (a : \Actor) \to (u : \Undergoer) \to \CulOrAtel\ a\ u \\
\pop'\ a\ (\B , \var{und}) \triangleq \textsf{pop}^\B\ a\ \var{und} \\
\pop'\ a\ (\U , \var{und}) \triangleq \pop\ a\ (\U,\var{und})
\end{gather*}
To see that each clause of $\pop'$ type-checks, we must unfold the definition of $\CulOrAtel\ a\ u$
according to the boundedness of the undergoer.

The type of $\pop'$ fully captures the behavior of \emph{pop} with respect to its telicity and
culminativity. Thus rather than postulating $\pop$, $\popped$, and $\textsf{popC}$, we could have
equivalently and more directly modeled \emph{pop} (\emph{eat}, etc.) by a function $(a : \Actor) \to
(u : \Undergoer) \to \CulOrAtel\ a\ u$.
\end{example}

\begin{remark}
There are also examples in English of telic but non-culminating events. The resulting state of the
telic event \emph{John cleaned the table} can be described by \emph{The table is clean}, but it is
not necessarily the case that \emph{John cleaned the table} $\Rightarrow$ \emph{The table is clean}. Such verbs are assigned the following type in our framework:
\[
(a : \Actor)\to (\var{u} : \Undergoer)\to \Evt^{\fst(u)}(a,\snd(\var{u}))
\]

As we discussed in Section~\ref{sec:bg-telicity}, there are also ambiguous verbs. The word
\emph{wipe} has two readings, one involving merely moving a cloth back and forth on some surface,
and the other involving an actual attempt to make the surface clean. The latter reading, like
\emph{clean}, is telic but non-culminating when its undergoer is bounded, and atelic when its
undergoer is unbounded. The former reading does not involve an approach towards any goal and so
exhibits no alternations in the telicity of the resulting event; it is therefore out of scope of our
investigation, as mentioned in Section~\ref{sec:Relationship between telicity and undergoers}.
\end{remark}


\begin{example}
Our framework can be also used to model kinds of verbs that do not exist in English. For example, as
discussed in \cite{kovalev2024modeling}, the so-called prototypical perfective verb forms in Russian
can only take bounded undergoers as arguments, giving rise to telic events which are moreover
necessarily culminating. (If their undergoer is a bare noun, then it receives a bounded reading.) In
our framework, we can assign such verb forms the type:
\[
  (a : \Actor)\to (\var{und} : \Undergoer^\B)\to \Cul(a,\var{und})
\]
\end{example}

\subsection{Adverbial modification}\label{sec:Adverbial modification}

Treating events as the types of their occurrences makes it easy to account for adverbial
modification in a parallel fashion to adjectival modification (Section~\ref{sec:Adjectival
modification}). We will be concerned only with manner adverbs; see \cite{Chatzikyriakidis2014,
2017_Chatzikyriakidis, 2020_Chatzikyriakidis_BOOK} for a discussion of dependently-typed treatments
of other kinds of adverbs, as well as other treatments of manner adverbs. We note that our treatment
below is in the spirit of \citet[359]{2015_Ranta_CONF}, although that work is not concerned with
dependent event types; rather unlike our treatment, \cite{2015_Ranta_CONF} models events as proof
objects of propositions.

Just as adjectives give rise to predicates over entities---instances of some noun phrase---we model
adverbs as predicates over \emph{occurrences}, or instances of some \emph{event}. Occurrences are
the equivalent in the verbal domain of the nominal domain's entities
(Section~\ref{sec:ProperNouns}). Thus far we have only had occasion to discuss occurrences
$\var{occ}:\El_\Evt(\var{evt})$ of a fixed event $\var{evt}:\Evt^b(a,\var{und})$, but we can define
the type of all occurrences $o:\Occ$ as pairs of events $e:\Evt$ with an instance of that event. (In
the following rule, recall that $e:\Evt$ is of the form $(a,(u,\var{evt}))$ where
$\var{evt}:\Evt^{\fst(u)}(a,\snd(u))$.)
\[
  \inferrule
  { }
  {\Gamma\vdash \Occ \triangleq \textstyle\sum_{e:\Evt} \El_{\Evt}(\snd(\snd(e))) \type}
\]

To mirror our construction of adjectival modification (Section~\ref{sec:Adjectival modification}),
we require that for any predicate $P$ over occurrences of some event $\var{evt}$, the ``subset of
occurrences of $\var{evt}$ satisfying $P$,'' $\sumEvt_{\var{occ}:\var{evt}} P$, is again an event
with the same actor and undergoer. That is, analogously to the rules for $\NP^b$ in
Section~\ref{sec:UniversesOfNP}, we require that the types $\Evt^b(a,\var{und})$ are closed under
$\Sigma$-types of predicates over $\El_\Evt(\var{evt})$:
\begin{mathpar}
  \inferrule
  {\Gamma\vdash b:\Boundedness \\
   \Gamma\vdash a:\Actor \\
   \Gamma\vdash \var{und}:\Undergoer^b \\\\
   \Gamma\vdash \var{evt}:\Evt^b(a,\var{und}) \\
   \Gamma,\var{occ}:\El_\Evt(\var{evt})\vdash P : \Prop}
  {\Gamma\vdash \sumEvt_{\var{occ}:\var{evt}} P : \Evt^b(a,\var{und})}
  \and
  \inferrule
  {\Gamma\vdash b:\Boundedness \\
   \Gamma\vdash a:\Actor \\
   \Gamma\vdash \var{und}:\Undergoer^b \\\\
   \Gamma\vdash \var{evt}:\Evt^b(a,\var{und}) \\
   \Gamma,\var{occ}:\El_\Evt(\var{evt})\vdash P : \Prop}
  {\Gamma\vdash \El_{\Evt}\left(\sumEvt_{\var{occ}:\var{evt}} P\right) =
   \textstyle\sum_{\var{occ}:\El_\Evt(\var{evt})} \Prff(P) \type}
\end{mathpar}

We can then model adverbs using predicates over $\Occ$. For example, using the predicate
$\quick:\Occ\to\Prop$ which holds for every \emph{quick} occurrence, we can represent an event
happening \emph{quickly}.

\begin{example}\label{ex:john ate apples quickly}
Let's illustrate how to represent the event \emph{John ate apples quickly}. As in
Example~\ref{ex:eat(john,apple) occurrences of events}, we have a family of events $\eat$ for any
actor and undergoer,
\[
  \inferrule
  { }
  {\Gamma\vdash \eat: (a:\Actor)\to (u:\Undergoer)\to \Evt^{\fst(u)}(a,\snd(u))}
\]
which we can apply to the actor \emph{John} and unbounded undergoer \emph{apple} to obtain the term
$\textsf{jaa} : \Evt^\U(\john_\Actor, \apple_\Undergoer)$ representing the event \emph{John ate
apples}.
\begin{gather*}
  \john_\Actor \triangleq \actTm_\Entity((\U,\human),\john) : \Actor \\
  \apple_\Undergoer \triangleq \undTm_\NP(\apple) : \Undergoer^\U \\
  \textsf{jaa} \triangleq \eat\ \john_\Actor\ (\U , \apple_\Undergoer)
  : \Evt^\U(\john_\Actor, \apple_\Undergoer)
\end{gather*}

We further add the predicate $\quick$ to our framework.
\[
  \inferrule
  { }
  {\Gamma\vdash \quick:\Occ\to\Prop}
\]
We can apply $\quick$ to any occurrence $o:\Occ$ whatsoever, but we can also restrict it by
precomposition to a predicate over occurrences $\var{occ}:\El_\Evt(\textsf{jaa})$ of the event
\emph{John ate apples}:
\[
  \lambda\var{occ}.\quick\ ((\john_\Actor,(\apple_\Undergoer,\textsf{jaa})),\var{occ})
  : \El_\Evt(\textsf{jaa})\to\Prop
\]

Using that observation, we model \emph{John ate apples quickly} as the event corresponding to the
subset of occurrences of \emph{John ate apples} that were \emph{quick}:
\[
  \textsf{jaaQuickly} \triangleq
  \sumEvt_{\var{occ}:\textsf{jaa}}
  \quick\ ((\john_\Actor,(\apple_\Undergoer,\textsf{jaa})),\var{occ})
  : \Evt^\U(\john_\Actor, \apple_\Undergoer)
\]

An occurrence of \emph{John ate apples quickly} is a term of type $\El_\Evt(\textsf{jaaQuickly})$.
By
\[
  \El_\Evt(\textsf{jaaQuickly}) =
  \textstyle\sum_{\var{occ}:\El_\Evt(\textsf{jaa})}
  \Prff(\quick\ ((\john_\Actor,(\apple_\Undergoer,\textsf{jaa})),\var{occ})) \type
\]
such terms are pairs of an occurrence $\var{occ}:\El_\Evt(\textsf{jaa})$ of \emph{John ate apples}
with a witness $\var{prf} : \Prff(\quick\
((\john_\Actor,(\apple_\Undergoer,\textsf{jaa})),\var{occ}))$ that $\var{occ}$ is \emph{quick}. In
particular, we can trivially prove the entailment \emph{John ate apples quickly} $\Rightarrow$
\emph{John ate apples}:
\[
\fst : \El_\Evt(\textsf{jaaQuickly}) \to \El_\Evt(\textsf{jaa})
\]
\end{example}

\subsection{Further entailments}\label{sec:Entailments}

We have already discussed several kinds of entailments: from transitive to intransitive sentences
(Examples~\ref{ex:john ate} and \ref{ex:pop balloon}), from culminating sentences to their resulting
states (Section~\ref{sec:culminating}), and from adverbially-modified sentences to
non-adverbially-modified sentences (Example~\ref{ex:john ate apples quickly}). All of these
entailments were accounted for by construction, but there are some other kinds of entailments that
our theory does not yet capture. In this section we consider two additional principles that each
account for a class of entailments concerning the undergoer of an event.

The first class of entailments is that the sentences describing telic events must entail the
sentences describing the corresponding atelic events. For example, \emph{John ate three apples}
should entail \emph{John ate apples}; that is, we should be able to construct a term of type
\[
  \El_\Evt(\eat\ \john_\Actor\ (\B,\undTm_\NP(\textsf{threeApples})))\to
  \El_\Evt(\eat\ \john_\Actor\ (\U,\undTm_\NP(\apple)))
  \tag{$\dagger$}
\]
where $\textsf{threeApples} \triangleq \AmountOf(\apple,\quantity,\natunits, 3):\NP^\B$ as in
Example~\ref{ex:john ate apples telicity}.

The general principle is as follows: for any event that occurs for actor $a$ and undergoer
$(\B,\undTm_\NP(\AmountOf(\var{np},d,u,m)))$, that same event should also occur with actor $a$ and
undergoer $(\U,\undTm_\NP(\var{np}))$. We add the following rule to our framework:
\[
  \inferrule{
    \Gamma \vdash d : \Degree \\
    \Gamma \vdash u : \Units(d) \\
    \Gamma \vdash m : \Nat \\\\
    \Gamma \vdash a : \Actor \\
    \Gamma \vdash f : (u : \Undergoer)\to\Evt^{\fst(u)}(a,\snd(u)) \\
    \Gamma \vdash \var{np}:\NP^\U \\
    \Gamma \vdash \var{occ} : \El_\Evt(f\ (\B , \undTm_\NP(\AmountOf(\var{np},d,u,m))))
  }{
    \Gamma \vdash \EvtAmtIsNP(f,\var{occ}) : \El_\Evt(f\ (\U,\undTm_\NP(\var{np})))
  }
\]

This rule has one subtlety worth noting: in the premise $f$ we see that the event (really, family of
events) must be applicable to any undergoer, bounded or unbounded, because the premise involves a
bounded undergoer and the conclusion a different, unbounded undergoer.

It is straightforward to check that $\lambda\var{occ}.\EvtAmtIsNP(\eat\ \john_\Actor,\var{occ})$ has
type $(\dagger)$ and thus establishes the entailment \emph{John ate three apples} $\Rightarrow$
\emph{John ate apples}.

The second class of entailments involves changing an undergoer from an entity to a bounded noun
phrase: \emph{Tom ate Jerry and Mickey} should entail \emph{Tom ate two mice}, i.e.
\[
\begin{gathered}
  \El_\Evt(\eat\ \tom_\Actor\ (\B,\undTm_\Entity((\B,\textsf{twoMice}),\textsf{jerryAndMickey})))\to
  \\
  \El_\Evt(\eat\ \tom_\Actor\ (\B,\undTm_\NP(\textsf{twoMice})))
\end{gathered}
\tag{$\ddagger$}
\]
assuming that we have terms
\begin{gather*}
\tom_\Actor:\Actor \\
\mouse : \NP^\U \\
\textsf{twoMice} \triangleq \AmountOf(\mouse,\quantity,\natunits,2) : \NP^\B \\
\textsf{jerryAndMickey} : \El_\NP^\B(\textsf{twoMice})
\end{gather*}

The general principle is expressed by the following rule, which we also add to our framework. Note
that the family of events $f$ need only be applicable to any bounded undergoer.
\[
  \inferrule{
    \Gamma \vdash d : \Degree \\
    \Gamma \vdash u : \Units(d) \\
    \Gamma \vdash m : \Nat \\\\
    \Gamma \vdash a : \Actor \\
    \Gamma \vdash f : (\var{und} : \Undergoer^\B)\to\Evt^\B(a,\var{und}) \\
    \Gamma \vdash \var{np}:\NP^\U \\
    \Gamma \vdash p : \El_\NP^\B(\AmountOf(\var{np},d,u,m)) \\
    \Gamma \vdash \var{occ} : \El_\Evt(f\ \undTm_\Entity((\B , \AmountOf(\var{np},d,u,m)),p))
  }{
    \Gamma \vdash \EvtEntIsNP(f,p,\var{occ})
    : \El_\Evt(f\ \undTm_\NP(\AmountOf(\var{np},d,u,m)))
  }
\]
Using the above rule, we can construct a term of type $(\ddagger)$ as follows:
\[
\lambda\var{occ}.
\EvtAmtIsNP\left(\left(\lambda\var{und}.\eat\ \tom_\Actor\ (\B,\var{und})\right),
\textsf{jerryAndMickey},\var{occ}\right)
\]
thereby establishing the entailment \emph{Tom ate Jerry and Mickey} $\Rightarrow$ \emph{Tom ate two
mice}.

%% file: sections/agda.tex
All of the framework rules and examples in this paper have been formalized in the Agda proof
assistant \citep{Agda}; our code is available at \url{https://doi.org/10.5281/zenodo.15602618}. Our
Agda formalization confirms that the rules in our framework are syntactically well-formed, and that
all our example definitions and entailments follow from the framework as described in the paper.

Our development consists of 637 lines of Agda code, including blank lines and comments, and is fully
self-contained. The code can be divided into the following components:
\begin{itemize}
\item Basic imports, definition of $\Prop$, and function extensionality: 30 lines.
\item Rules from Section~\ref{sec:Nominal} (nominal framework): 130 lines.
\item Examples from Section~\ref{sec:Nominal}: 164 lines.
\item Rules from Section~\ref{sec:Verbal} (verbal framework): 113 lines.
\item Examples from Section~\ref{sec:Verbal}: 200 lines.
\end{itemize}

We encode the framework rules into Agda using a combination of postulates, rewrite rules along
postulated equations, and ordinary data and function definitions. For example,

\ExecuteMetaData[agda/latex/Section5.tex]{NP}

\noindent
postulates two functions $\AgdaPostulate{NP}$ and $\AgdaPostulate{ElNP}$ whose codomains are Agda's
lowest type universe $\AgdaPrimitive{Set}$.\footnote{In Agda, $\{\AgdaBound{b} :
\AgdaDatatype{Bd}\}\to\cdots$ is the notation for a $\Pi$-type whose argument is left implicit by
default.} This has the same effect as the following two rules from Section~\ref{sec:UniversesOfNP}:
\begin{mathpar}
  \inferrule
  {\Gamma \vdash b:\Boundedness}
  {\Gamma\vdash \NP^b \type}
  \and
  \inferrule
  {\Gamma\vdash b:\Boundedness \\
   \Gamma\vdash \var{np}:\NP^b}
  {\Gamma\vdash \El_{\NP}^b(\var{np}) \type}
\end{mathpar}

Some of our framework rules express judgmental equalities, such as the second rule below:
\begin{mathpar}
  \inferrule
  {\Gamma\vdash b:\Boundedness \\
   \Gamma\vdash \var{np}:\NP^b \\
   \Gamma,p:\El_{\NP}^b(\var{np})\vdash P : \Prop}
  {\Gamma\vdash \sumNP_{p:\var{np}} P : \NP^b}
  \and
  \inferrule
  {\Gamma\vdash b:\Boundedness \\
   \Gamma\vdash \var{np}:\NP^b \\
   \Gamma,p:\El_{\NP}^b(\var{np})\vdash P : \Prop}
  {\Gamma\vdash \El_{\NP}^b\left(\sumNP_{p:\var{np}} P\right) =
   \textstyle\sum_{p:\El_{\NP}^b(\var{np})} \Prff(P) \type}
\end{mathpar}
We encode these equations by first postulating them as propositional equalities and then declaring
them as rewrite rules \citep{CockxTabareauWinterhalter21}, a feature of Agda which allows upgrading
some propositional equalities (satisfying certain conditions) to judgmental equalities.

\ExecuteMetaData[agda/latex/Section5.tex]{NPSigma}

We can then use the Agda version of our framework to model fragments of natural language by
postulating the relevant terms ($\AgdaPostulate{human}$, $\AgdaPostulate{john}$,
$\AgdaPostulate{eat}$, etc.) and defining terms corresponding to various events and entailments.
Revisiting Example~\ref{ex:john ate apples quickly} (\emph{John ate apples quickly}), we can define
the event \emph{John ate apples} as follows:

\ExecuteMetaData[agda/latex/Section5.tex]{johnAteApples}

We then define a function $\AgdaFunction{-ly}$ which adverbially modifies any event by pairing its
occurrences with a proof that they satisfy a given predicate $\Occ\to\Prop$.

\ExecuteMetaData[agda/latex/Section5.tex]{ly}

\noindent
In particular, applying $\AgdaFunction{-ly}~\AgdaFunction{quick}$ to $\AgdaFunction{johnAteApples}$
corresponds exactly to the definition of $\textsf{jaaQuickly}$ given in Example~\ref{ex:john ate
apples quickly}.

\ExecuteMetaData[agda/latex/Section5.tex]{johnAteApplesQuickly}

Agda then confirms that $\fst$ is a proof that \emph{John ate apples quickly} entails \emph{John ate
apples}.

\ExecuteMetaData[agda/latex/Section5.tex]{eg}

%% file: sections/future.tex
In this paper, we have developed a cross-linguistic framework within Martin-L\"of type theory for
analyzing event telicity and culminativity. Our framework consists of two parts. First, in the
nominal domain, we develop a compositional analysis of the boundedness of noun phrases and of the
internal structure of overtly bounded noun phrases. Then, in the verbal domain, we develop a
dependently-typed calculus for static and dynamic eventualities as a backdrop for defining telicity
and culminativity, and deriving associated entailments (in particular, that occurrences of
culminating events entail the associated resulting state). We illustrate the applicability of our
framework through a series of examples modeling English sentences; the first author's dissertation
additionally discusses Russian \citep{kovalev2024modeling}.

We have already discussed prior work at length in Section~\ref{sec:Introduction}; in this section we
close with some more detailed comparisons to related work, particularly to
\cites{2020_Corfield_BOOK} type-theoretic approach to Vendler's Aktionsart classes, and to
MTT-semantics \citep{2020_Chatzikyriakidis_BOOK} and its treatment of event types
\citep{2017_Luo_CONF} and noun phrases \citep{2013_Chatzikyriakidis_CONF}. We also indicate some
possible directions for future work.

\paragraph*{Martin-L\"of type theory and its semantics}

As previously discussed, many aspects of our framework are inspired by MTT-semantics; perhaps our
largest departure from MTT-semantics is that we build atop Martin-L\"of type theory
\citep{MartinLof75itt} rather than the Unifying Theory of Dependent Types (UTT)
\citep{1994_Luo_BOOK} extended with coercive subtyping and other features. We have opted to use
Martin-L\"of type theory because it is more theoretically parsimonous and more widely used, studied,
and implemented than UTT with coercive subtyping; the former is, roughly speaking, the core language
of the Agda proof assistant.

As an example of this parsimony, Section~\ref{sec:Subtyping in the nominal domain} demonstrates that
we can model subtyping relationships such as \emph{every man can be regarded as a human} by positing
a family of types $\isA(\var{np},\var{np}')$ along with a function $\El_\isA(-) :
\isA(\var{np},\var{np}')\to \El_{\NP}^b(\var{np})\to \El_{\NP}^{b'}(\var{np}')$ and an axiom
$\IARespectsIsA$ (in Section~\ref{sec:Adjectival modification}) stating that intersective adjectives
lift along $\isA$ relationships. In MTT-semantics, linguistic subtyping is modeled by coercive
subtyping relations such as $\man\leq\human$, which in turn are governed by rules in the ambient
type theory which insert implicit coercions at function applications \citep{2009_Luo_CONF}. Although
coercive subtyping is conceptually more direct, it complicates the metatheory and semantics of type
theory \citep{Soloviev2001} in comparison to our explicit inclusion functions. We do not claim that
all of MTT-semantics can be reconstructed in Martin-L\"of type theory, but at least for the purposes
of this paper we were able to avoid using any non-standard type-theoretic features.

In particular, building on a standard type theory means that our framework has a straightforward
set-theoretic semantics with respect to which we can validate the entailment relations asserted
throughout this paper. \citet{LuoBOTH2019} has previously argued that, in addition to being
evidently proof-theoretic, MTT-semantics also provides a model-theoretic semantics ``not [in the
sense of being] given a set-theoretic semantics'' but rather that ``an MTT can be employed as a
meaning-carrying language to give the model-theoretic semantics to Natural Language.''

We agree with the latter point but wish to emphasize that our framework \emph{also} has a
set-theoretic semantics, obtained by extending the set-theoretic semantics of Martin-L\"of type
theory as explicated e.g.\ by \citet{Hofmann97} or \citet[Section 3.5]{Angiuli2024}. This semantics
interprets closed types $\cdot\vdash A\type$ as sets $\semantics{A}$, closed terms $\cdot\vdash a:A$
as elements of $\semantics{A}$, dependent families of types $x:A\vdash B(x)\type$ as
$\semantics{A}$-indexed families of sets $\semantics{B}(x)$ for $x\in\semantics{A}$, and dependent
families of terms $x:A\vdash b(x) : B(x)$ as elements of the $\semantics{A}$-indexed product of sets
$\prod_{x\in\semantics{A}} \semantics{B}(x)$. This semantics moreover interprets $\Pi$-types as
set-indexed products of sets, $\Sigma$-types as set-indexed disjoint unions of sets, and intensional
identity types $\mathbf{Id}_A(a,a')$ as subsingleton sets $\{\star \mid a = a'\}$.

As for the additional rules stipulated by our framework, a set-theoretic semantics of our framework
in the case of e.g.\ English would take $\semantics{\Boundedness}$ to be the set $\{\B,\U\}$;
$\semantics{\NP^\B}$ (resp., $\semantics{\NP^\U}$) to be the set of bounded (resp., unbounded)
English noun phrases under consideration; for a given English noun phrase
$\textit{human}\in\semantics{\NP^\U}$ under consideration, takes
$\semantics{\El_\NP^\U}(\textit{human})$ to be the set of instances of \emph{human} under
consideration; takes $\semantics{\isA}(\var{np},\var{np}')$ to be $\{\star \mid
\semantics{\El_\NP^b}(\var{np}) \subseteq \semantics{\El_\NP^{b'}}(\var{np}') \}$; and so forth.

These sets validate the rules of our framework---at least insofar as our framework correctly models
the English language---and provide a reference semantics with respect to which we can check the
entailments predicted by our framework. Describing and validating this semantics is one clear avenue
for future work.

We stress in particular that the formal semantics of natural language based on Martin-L\"of type
theory is no less model-theoretic than traditional Montague semantics based on simple type theory,
nor the semantics based on Type Theory with Records \citep{2005_Cooper, Cooper2011,
2015_Cooper_CONF, 2017_Cooper_CONF, 2023_Cooper_BOOK} as suggested e.g.\ by
\citet[116--117]{Sutton2024}. Nor does working inside type theory commit semanticists to
constructivity; in set-theoretic models, sentences can still be understood as being simply true or
false. For instance, in the context of Example~\ref{ex:black cats formal}, one can create models in
which $\tom:\El_{\NP}^\U(\cat)$ is a black cat without stipulating a term $\tomIsBlack :
\Prff(\El_{\IA}(\blk)\ ((\U,\cat),\tom))$ in the syntax.

\paragraph*{Event semantics and dependent event types}

Various forms of event semantics or event calculi have long been used in both linguistics and
philosophy for natural language semantics \citep{Davidson1967-DAVTLF, Castaneda1967,
Higginbotham1985, Higginbotham2000, 1990_Parsons_BOOK, Parsons2000}\footnote{A linguistic survey can
be found in \cite{Maienborn2011, Maienborn2019}; a survey from the perspective of the philosophy of
language can be found in \cite{Williams2021}.} and for reasoning about action and change in
artificial intelligence \citep{Kowalski1986, Shanahan1995, Shanahan1997, Shanahan1999, Miller2002,
Mueller2008, Mueller2014}.

A type-theoretic approach to event semantics goes back at least as far as \cite{2015_Ranta_CONF},
who was motivated by the non-compositional nature of standard event semantics: as discussed in
\citet[359]{2015_Ranta_CONF}, the Davidsonian representation of the sentence \emph{John buttered the
toast in the bathroom with a knife at midnight}, given in \ref{Jones Davidsonian}, is not
compositional because \emph{John buttered the toast} is not a constituent of the final proposition.
This problem can be resolved through the use of $\Sigma$-types, as demonstrated in \ref{Jones
Sigma}.

\ex. 
\a. $(\exists e)(\butter(\john,\toast,e)\land \inbathroom(e)\land \withknife(e)\land \atmidnight(e))$ \label{Jones Davidsonian}
\b. $\textstyle\sum_{e:\butter(\john,\toast)} \inbathroom(e)\land \withknife(e)\land \atmidnight(e)$ \label{Jones Sigma}

Here \citet{2015_Ranta_CONF} treats $\butter(\john,\toast)$ as a proposition, and events
$e:\butter(\john,\toast)$ as proofs of that proposition. A similar approach is taken by
\citet[19]{2023_Cooper_BOOK}, in which $e: \run(a)$ means that $e$ is an event in which the
individual $a$ is running.

Our approach is compositional in the sense of \cite{2015_Ranta_CONF} but differs from his approach
in two respects. First, we represent events as types rather than propositions; for us
$\butter(\john,\toast)$ is itself an event whose terms (of which there may be many) are occurrences
of that event. Second, since we keep track of actors, undergoers, and telicity,
$\butter(\john,\toast)$ is not just a bare type but an element of the collection (universe) of
events $\Tel(\john,\toast)$ (assuming that $\toast$ represents a bounded noun phrase).

Our way of keeping track of actors and undergoers is inspired by \cites{2017_Luo_CONF} dependent
event types, which incorporate the so-called neo-Davidsonian event semantics
\citep{1990_Parsons_BOOK} into the framework of MTT-semantics. \citet{2017_Luo_CONF} introduce event
types with a fixed agent and/or patient (or more generally actor and undergoer, in our terminology);
for example, $Evt_{AP}(a,p)$ (resp., $Evt_A(a)$) is the type of events with agent $a$ and patient
$p$ (resp., with agent $a$), and these are related by coercive subtyping relationships such as
$Evt_{AP}(a,p) \leq Evt_A(a)$.

In contrast, we have a single primitive type $\Evt^b(a,\var{und})$ of events with a specified actor
and undergoer, in terms of which the less-specified types of events (and explicit coercion functions
between them) can be defined using $\Sigma$-types. Unlike \citet{2017_Luo_CONF} we also explicitly
distinguish between events that have no actor and/or undergoer, which we represent with dummy
arguments, and events with an unspecified actor and/or undergoer.


As an illustration of dependent event types in MTT-semantics, \ref{ex:barkWOevents} is the type of
\emph{bark} in MTT-semantics without event types, and \ref{ex:barkWevents} is its type in the
presence of event types.

\ex. 
\a. $\bark: \dog \to \Prop$\label{ex:barkWOevents} \hfill MTT-semantics without events
\b. $\bark : (x:\dog)\to Evt_A(x) \rightarrow \Prop$\label{ex:barkWevents} \hfill MTT-semantics with events

A sentence like \emph{Fido barks} is then modeled as \ref{fidobarks}.

\ex. $(\exists e:Evt_A(\textsf{fido}))(\bark(\textsf{fido},e))$ \label{fidobarks}

\relax

The main application of dependent event types discussed in \cite{2017_Luo_CONF} is their elegant
solution to the event quantification problem \citep{2011_Winter_CONF, 2015_Groote_CONF,
2015_Champollion}, roughly, that e.g.\ the sentence \emph{No dog barks} should mean ``There's no dog
$d$ for which there exists a corresponding barking event whose agent is $d$'' as opposed to ``There
exists an event in which  no dog is barking.''

Like traditional (non-dependent) event semantics, however, their approach treats events as variables
and represents every sentence as starting with an existential quantifier over events, as in
\ref{fidobarks}. We find this quite unnatural for several reasons. First, it requires a stipulative
post-compositional operation to bind the event variable. Second, having every sentential
representation begin with an existential quantification over events unnecessarily complicates
sentential representations. Third, there is no reasonable way to interpret the event variable. The
original motivation for such variables in linguistic semantics was to allow for adverbial
modification, but we have already demonstrated an alternate solution in Section~\ref{sec:Adverbial
modification}.

In summary, the dependent event types in our framework are similar to those of \citet{2017_Luo_CONF}
except that we treat events as types instead of variables, which not only avoids the issues
described above but also avoids the event quantification problem altogether.

\paragraph*{More detailed event structure}

Our work focuses on the inner aspectual properties of telicity, culminativity, and to some extent,
staticity, but has ignored other interesting inner aspectual properties such as punctuality or
durativity (which we use as the antonym of punctuality).\footnote{For example, the event of a
balloon popping is punctual, whereas the event of ice melting is durative.} If one considers
culminativity, punctuality, and staticity, one arrives at Vendler's ontology of Aktionsart classes
\citep{1957_Vendler, 1967_Vendler_BOOK}: states, activities, achievements, and
accomplishments.\footnote{If one distinguishes between telicity and culminativity as we do, then one
should add a class for non-culminating accomplishments. Additionally, there is a standard
non-Vendlerian Aktionsart class of semelfactives \citep{1997_Smith_BOOK}.} Our work explicitly
covers states; it partially covers activities---all atelic events are activities, but our framework
doesn't cover activities such as \emph{push the cart}; and we have collapsed achievements and
accomplishments (either culminating or non-culminating ones) into telic events, because we do not
distinguish durativity and punctuality. One interesting avenue for future work is to extend our
framework with the punctuality--durativity distinction.

Some work on this has already has been done by \citet{2020_Corfield_BOOK}, who uses $\Sigma$-types
to formalize \cites{Moens1988} notion of event nucleus, which is described as ``an association of a
goal event, or culmination, with a preparatory process by which it is accomplished, and a consequent
state, which ensues'' \citep[15]{Moens1988}. \cite{2020_Corfield_BOOK} introduces the types of
states, activities, achievements, and accomplishments, the first three being primitive and the
fourth defined as in \ref{Corfield accomplishment}. The event nucleus is then defined either as in
\ref{Corfield event nucleus 1} or as in \ref{Corfield event nucleus 2},
where $\Culminate$ and $\Consequent$ are taken as
primitive.

\ex. 
\a. $\Accomplishment := \sum \limits_{\substack{x : \Activity \\ y : \Achievement}} \Culminate(x, y)
$ \label{Corfield accomplishment}
\b. $\EventNucleus := \sum \limits_{\substack{x : \Activity \\ y : \Achievement \\ z : \State}} (\Culminate(x, y) \land \Consequent(y, z))$ \label{Corfield event nucleus 1}
\b. $\EventNucleus := \sum \limits_{\substack{w : \Accomplishment \\ z : \State}} \Consequent(\fst(\snd(w)), z)
$ \label{Corfield event nucleus 2}

\relax

Relative to \citet{2020_Corfield_BOOK}, our work is both an improvement and an oversimplification.
It is an improvement in the sense that we account for verbal arguments and define what it means for
an event to culminate, in terms of a resulting state. On the other hand, as mentioned above, we do
not attempt to model the distinction between accomplishments and achievements. Further work is
needed to combine the two approaches in a reasonable way.

\paragraph*{Structure of noun phrases}

Surprisingly, the internal structure of noun phrases (e.g., the distinction between \emph{apples},
\emph{two apples}, and \emph{two kilograms of apples}) has received little to no attention in the
literature on type-theoretic semantics of natural language.

Work on copredication \citep{Chatzikyriakidis2017, 2018_Chatzikyriakidis} has proposed treating
numerical quantifiers in the spirit of generalized quantifier theory, i.e., taking nouns and verbs
as arguments. We deviate from this approach for several reasons. First, this treatment relies on
modeling collections of common nouns as setoids, which we have avoided. Second, this treatment of
numerical quantifiers makes the structure of sentences like \emph{John ate three apples} very
different from that of sentences like \emph{John ate apples}, which introduces a non-uniformity and
may complicate entailments such as the former sentence entailing the latter.

Mass nouns (such as \emph{water}) as well as the so-called pseudo-partitive constructions (such as
\emph{a glass of water} or \emph{three kilograms of apples}) seem to have barely received any
attention in the literature on MTT-semantics. We are only aware of a paragraph in
\cite{2012_Luo_CONF}, which proposes representing \emph{John drank a glass of water} as in
\ref{drink a glass of water Luo}, which treats \emph{glass of water} as one atomic unit.
\citet{2012_Luo_CONF} also proposes that mass nouns used without a measure word are underspecified
and can thus be modeled with overloading in MTT-semantics, but no further details are given.

\ex. $\exists w : \llbracket \text{glass of water} \rrbracket . \llbracket \text{drink} \rrbracket(j, w)$ \label{drink a glass of water Luo} \hfill \cite{2012_Luo_CONF}


In this paper we account for the internal structure of overtly bounded noun phrases using degrees
and units. A similar but less refined treatment can be found in \cite{2013_Chatzikyriakidis_CONF},
in which the type of \emph{John and Mary} is $\Vect(\human,2)$ (in contrast to our
$\AmountOf(\human,\quantity, \natunits, 2)$), and coercive subtyping relationships such as $\human
\leq \Vect(\human,1)$ allow considering terms of type $\human$ as having type $\Vect(\human,1)$.


\paragraph*{Connection to mereological approaches}

Our framework also contains an $\oplus$ operation for producing sums of instances. This idea goes
back at least as far as \cite{1983_Link}, where pluralities are modeled as ``sums of individuals''
in the sense of classical extensional mereology \citep{2016_Champollion_Chapter, Pietruszczak2020,
2021_Cotnoir_Varzi_BOOK}.\footnote{See \cite{2022_Champollion_CONF} for an overview of
\cite{1983_Link} and its connection to more recent developments.}

Mereological sums have also proved useful in the study of the properties of distributivity (e.g.\
\emph{John and Mary ran} entails \emph{John ran} and \emph{Mary ran}) and collectivity (e.g.\
\emph{John and Mary met} does not entail \emph{John met} and \emph{Mary met}); see
\cite{Champollion2019} for an overview. To the best of our knowledge there are no studies of
mereological approaches to natural language semantics in dependent type theory. There is only one
mention of collective verbs such as \emph{meet} in \cite{2013_Chatzikyriakidis_CONF}, in which
\emph{meet} would have the following type:

\ex. $\meet: (n:\Nat)\to \Vect(\human, n+2)\to \Prop$ \label{meet}

\relax

Some interesting future research directions would be to develop a system within dependent type
theory which treats the distributive-collective opposition, as well as the related phenomena of
cumulativity \citep{2020_Champolli} and reciprocity \citep{Winter2018}. More modestly, one might add
further axioms to our framework which make $\oplus$ a join semilattice operator.

\paragraph*{Adjectival and adverbial modification}

We have only discussed modification of noun phrases by intersective adjectives, but not by other
kinds of adjectives or relative clauses. With respect to boundedness---our primary concern in the
nominal domain---these other kinds of modifiers seem to, like intersective adjectives, preserve the
boundedness of noun phrases.

Of course, these modifiers will behave differently in other respects and are worth consideration.
Subsective, privative, non-committal, gradable, and multidimensional adjectives have been studied in
MTT-semantics \citep{2017_Chatzikyriakidis, 2020_Chatzikyriakidis_BOOK, 2022_Chatzikyriakidis_CONF}.
We expect that our framework can account for subsective and privative adjectives without additional
machinery; gradable and multidimensional adjectives will require some additional work on degrees;
and non-committal adjectives will require stepping into the modal territory.

In the verbal domain, we have only considered adverbial modification of events. Other kinds of
adverbs include manner adverbs (e.g., \emph{John wrote illegibly}, where it is not the event that is
illegible, but rather the manner in which John wrote), actor-oriented verbs (e.g., \emph{Clumsily,
John dropped his cup of coffee}, where it is John's act of dropping his cup of coffee that was
clumsy\footnote{But note that the adverb in \emph{John clumsily dropped his cup of coffee} is
ambiguous between a manner adverb and an actor-oriented adverb.}), speech act adverbs (as in
\emph{Frankly, my dear, I don't give a damn}), and intensional adverbs (as in \emph{Oedipus
allegedly married Jocaste}), the latter of which require modalities as in
\cite{2020_Chatzikyriakidis_BOOK}. The other types of adverbs listed above are accounted for in
\cite{2020_Chatzikyriakidis_BOOK} by adding more dependency to event types, which can be adapted to
our framework. For example, to account for manner adverbs, one can introduce the type of events with
a given actor, undergoer, and manner.

\paragraph*{Extensions to other sources of telicity}

In this paper we have restricted ourselves to considering verbs whose lexical semantics involves the
aim of achieving a certain goal---thereby ignoring sentences like \emph{John pushed a cart}---and
have only considered sentences in which telicity comes from verbal arguments rather than adjuncts
(as in \emph{John drove a car to Bloomington}) or resultative complements (as in \emph{John wiped
the table clean}). Future work is needed to extend our framework to account for these other kinds of
telicity.